\documentclass[conference]{IEEEtran}
\usepackage{times}

\usepackage[numbers]{natbib}
\usepackage{multicol}
\usepackage[bookmarks=true]{hyperref}

\usepackage{booktabs}       
\usepackage{tabularx}
\usepackage{mathtools}
\usepackage[vlined,ruled,linesnumbered,procnumbered]{algorithm2e}
\usepackage{microtype}      
\usepackage{url}
\usepackage{graphicx}
\usepackage{stmaryrd}
\usepackage{multirow}
\usepackage[table,dvipsnames]{xcolor}

\usepackage{verbatim}
\usepackage{amsthm}
\usepackage{siunitx}
\usepackage{cleveref}
\usepackage{tikz}
\usetikzlibrary{fit}
\usepackage{amssymb}
\usepackage{nicematrix}
\usepackage{mathptmx}
\DeclareMathAlphabet{\mathcal}{OMS}{cmsy}{m}{n}

\SetProcNameSty{textsc}
\SetFuncSty{textsc}

\SetCommentSty{mycommfont}

\usepackage{bm}
\usepackage{mathrsfs}

\newcommand{\RR}{\mathbb{R}}

\newcommand{\NN}{\mathbb{N}}

\newcommand{\calK}{\mathcal{K}}

\newcommand{\calV}{\mathcal{V}}

\newcommand{\scrL}{\mathscr{L}}

\newcommand{\subjto}{\mathrm{s.t.}\ }

\newcommand{\defeq}{\stackrel{\mathclap{\tiny\mathrm{def}}}{=}}

\newcommand{\bmlam}{\bm{\lambda}}

\newcommand{\bmx}{\bm{x}}
\newcommand{\bmu}{\bm{u}}

\newcommand{\bmnu}{\bm{\nu}}

\newcommand{\calE}{\mathcal{E}}
\newcommand{\calI}{\mathcal{I}}
\newcommand{\calP}{\mathcal{P}}
\newcommand{\calO}{\mathcal{O}}

\renewcommand\leq{\leqslant}
\renewcommand\geq{\geqslant}

\definecolor{paleaqua}{rgb}{0.74, 0.83, 0.9}
\definecolor{pastelblue}{rgb}{0.68, 0.78, 0.81}

\newcommand{\symmat}[1]{\mathbb{S}_{#1}(\RR)}

\colorlet{lightApricot}{Apricot!93}
\colorlet{myblue}{blue!44!white}
\def\arrstyle{\color{blue!65!darkgray}}

\newtheorem{prop}{Proposition}

\newtheorem{remark}[prop]{Remark}

\begin{document}

\title{Parallel and Proximal Constrained Linear-Quadratic Methods %
    for Real-Time Nonlinear MPC}



%
\author{%
    \authorblockN{Wilson  Jallet\authorrefmark{1}\authorrefmark{2}\
    Ewen Dantec\authorrefmark{2}\
    Etienne Arlaud\authorrefmark{2}\ 
    Nicolas Mansard\authorrefmark{1}\authorrefmark{3}\
    Justin Carpentier\authorrefmark{2}}
    \authorblockA{\authorrefmark{1}LAAS-CNRS, University of Toulouse\\
    31400 Toulouse, France\\
    \{wjallet,nmansard\}@laas.fr}
    \authorblockA{\authorrefmark{2}Inria - Département d’Informatique de l’École Normale Supérieure,
PSL Research University\\
    75012 Paris, France\\
    \{wilson.jallet,ewen.dantec,etienne.arlaud,justin.carpentier\}@inria.fr}
    \authorblockA{\authorrefmark{3}ANITI, University of Toulouse\\
    Toulouse, France}
}

\maketitle

\begin{abstract}
Recent strides in nonlinear model predictive control (NMPC) underscore a dependence on numerical advancements to efficiently and accurately solve large-scale problems. Given the substantial number of variables characterizing typical whole-body optimal control (OC) problems —often numbering in the thousands— exploiting the sparse structure of the numerical problem becomes crucial to meet computational demands, typically in the range of a few milliseconds.
Addressing the linear-quadratic regulator (LQR) problem is a fundamental building block for computing Newton or Sequential Quadratic Programming (SQP) steps in direct optimal control methods. This paper concentrates on equality-constrained problems featuring implicit system dynamics and dual regularization, a characteristic of advanced interior-point or augmented Lagrangian solvers.
Here, we introduce a parallel algorithm for solving an LQR problem with dual regularization. Leveraging a rewriting of the LQR recursion through block elimination, we first enhanced the efficiency of the serial algorithm and then subsequently generalized it to handle parametric problems. This extension enables us to split decision variables and solve multiple subproblems concurrently.
Our algorithm is implemented in our nonlinear numerical optimal control library \textsc{aligator}\footnotemark[1].
It showcases improved performance over previous serial formulations and we validate its efficacy by deploying it in the model predictive control of a real quadruped robot.

\footnotetext[1]{\label{note1}Repository: \url{https://github.com/Simple-Robotics/aligator/}}

\end{abstract}

\IEEEpeerreviewmaketitle

\section{Introduction}
\label{sec:intro}

In this paper, we introduce a parallel algorithm to enhance the efficiency of model-predictive control (MPC) solvers~\cite{tassaSynthesisStabilizationComplex2012,dantecFirstOrderApproximation2022}.
The computational complexity of these solvers is a pivotal factor in numerical optimal control -- in particular, to allow their implementation on real hardware.
More specifically, we consider linear-quadratic (LQ) problems (i.e., with quadratic cost and linear constraints), which are a fundamental block of such iterative solvers. LQ is the standard form of subproblems in many direct methods derived from Newton's method~\cite{dunnEfficientDynamicProgramming1989}
such as sequential quadratic programming (SQP)~\cite{diehlFastDirectMultiple2006,giftthalerFamilyIterativeGaussNewton2018}, differential dynamic programming (DDP), and iterative LQR~\cite{jacobsonDifferentialDynamicProgramming1970,tassaSynthesisStabilizationComplex2012}.

In particular, the classical linear-quadratic regulator (LQR) is an LQ problem defined in terms of explicit linear dynamics and an unconstrained objective.
The most well-known method for solving the classical LQR is the Riccati recursion, which can be derived by dynamic programming~\cite{bertsekas2019reinforcement}.
Yet, direct equivalence with block-factorization methods has been drawn~\cite{wrightSolutionDiscretetimeOptimal1990} and recently exploited in robotics~\cite{jordanaStagewiseImplementationsSequential2023}.

Over the past decade, proposals have been given for the resolution of nonlinear equality-constrained problems.
Most solutions essentially extend the Riccati recursion approaches to properly account for equality constraints, by exploiting projection or nullspace approaches~\cite{giftthalerProjectionApproachEquality2017,laineEfficientComputationFeedback2019,vanroyeGeneralizationRiccatiRecursion2023} or augmented Lagrangian-based approaches~\cite{kazdadiEqualityConstrainedDifferential2021,howellALTROFastSolver2019}.
While solving the LQR is often a bottleneck in recent efficient optimal control solvers~\cite{ocs2,mastalliCrocoddylEfficientVersatile2020,frisonAlgorithmsMethodsHighPerformance}, most of them rely on sequential implementation without exploiting the parallelization capabilities of modern processing units.

Several methods were previously developed for the parallel solving of LQ problems:
\cite{dengHighlyParallelizableNewtontype2018} is based on a Gauss-Seidel modification of the Riccati backward sweep, ADMM schemes separating costs and constraints~\cite{stathopoulosHierarchicalTimesplittingApproach2013} and conjugate-gradient methods~\cite{adabagMPCGPURealTimeNonlinear2023} (efficient on GPUs), all with intrinsic approximate (iterative) convergence at linear rate.

On the other hand, 
Wright~\cite{wrightSolutionDiscretetimeOptimal1990,wrightPartitionedDynamicProgramming1991} looked at direct methods for solving linear-quadratic problems on parallel architectures. (first with a tailored banded matrix LU solver~\cite{wrightSolutionDiscretetimeOptimal1990}, then~\cite{wrightPartitionedDynamicProgramming1991} with a specialized method using dynamic programming after partitioning the problem, with an extension to active-set methods).
%
More recently, \citet{nielsenLogParallelAlgorithm2014} subdivide the LQR problem into subproblems with state-control linkage constraints; this approach involves computing nullspace matrices to handle infeasible subproblems. Then, \cite{nielsenParallelStructureExploiting2015} introduces a variant based on parameterizing each subproblem on the next subproblem value function's parameters.
\citet{laineParallelizingLQRComputation2019} suggest subdivision of the LQR problem into a set of subproblems with state linkage constraints at the endpoints; the linkage constraints' multipliers satisfy a global system of equations which is solved by least-squares.

In this paper, we propose a general direct solver for LQ problems with implicit dynamics and additional equality constraints, leveraging parameterization to formulate a parallel algorithm, a similar idea to \citet{nielsenParallelStructureExploiting2015,laineParallelizingLQRComputation2019}.
This novel algorithm is implemented in \textsf{C++}, and used as a backend in a nonlinear trajectory optimizer which handles both equality and inequality constraints using an augmented Lagrangian method.
Its effectiveness is demonstrated on various robotic benchmarks and by implementing an MPC scheme on a real quadruped robot.
This paper follows up from our prior work on augmented Lagrangian methods for numerical optimal control with implicit dynamics and constraints~\cite{jalletImplicitDifferentialDynamic2022,jalletConstrainedDifferentialDynamic2022}.

After recalling the equality-constrained LQR problem in \Cref{sec:elq_problem}, we first derive serial Riccati equations in the proximal setting in \Cref{sec:riccati}, for which we build a block-sparse factorization in \Cref{sec:blocksparse}.
This formulation is extended in \Cref{sec:parametric} to parametric LQ problems, which we finally use in \Cref{sec:parallel} to build a parallel algorithm and discuss it with respect to the literature. 
The application of our algorithm in proximal nonlinear trajectory optimization is described in \Cref{sec:nonlinear}, along with performance benchmarks and experiments in \Cref{sec:experiments}.

\subsection*{Notation}

In this paper, we denote, $\RR^{n\times m}$ the set of $n\times m$ real matrices, and $\symmat{n}$ the set of $n\times n$ symmetric real matrices.
We denote \mbox{$\llbracket a,b\rrbracket = \{a, a + 1,\ldots,b\}$} the interval of integers between two integers $a \leq b$.
We will use italic bold letters to denote tuples of vectors $\bm z = (z_0, \ldots, z_k)$ of possibly varying dimensions, and given an index set $\calI\subseteq \llbracket 0,k\rrbracket$, we denote by $\bm{z}_{\calI}$ the subset $(z_i)_{i\in\calI}$ (replacing $\calI$ by its intersection with $\llbracket 0,k\rrbracket$ when not contained in the former, by abuse of notation).
For linear systems $Hz + g = 0$, we will use the shorthand notation
\begin{equation*}
	\begin{bNiceArray}[first-row]{c|c}
		\RowStyle{\arrstyle}
		z \\
		H & g
	\end{bNiceArray}
\end{equation*}
for compactness.

\section{Equality-constrained linear-quadratic Problems}
\label{sec:elq_problem}

In this section, we will recall the equality-constrained linear-quadratic (LQ) problem, its optimality conditions, and proximal methods to solving it.

\subsection{Problem statement}

We consider the following equality-constrained LQ problem:
\begin{subequations}\label{eq:lqr:problem}
\begin{alignat}{3}
    \min_{\bmx, \bmu} &\ J(\bmx, \bmu) \defeq
    \sum_{t=0}^{N-1} \ell_t(x_t, u_t) + \ell_N(x_N) \\
    \subjto
    &A_tx_t + B_tu_t + E_tx_{t+1} + f_t = 0 \label{eq:lqr:dyn} \\
    &C_tx_t + D_tu_t + h_t = 0, \ t=0,\ldots,N-1 \\
    &C_Nx_N + h_N = 0 \\ \label{eq:lqr:init}
    &G_0x_0 + g_0 = 0
\end{alignat}
\end{subequations}
with the quadratic running and terminal cost functions
\begin{subequations}\label{eq:lqr:problem:costs}
\begin{align}
    \ell_t(x_t, u_t) &= \frac{1}{2}
    \begin{bmatrix}
        x_t \\ u_t
    \end{bmatrix}
    \begin{bmatrix}
        Q_t & S_t \\ S_t^\top & R_t
    \end{bmatrix}
    \begin{bmatrix}
        x_t \\ u_t
    \end{bmatrix}
    + q_t^\top x_t + r_t^\top u_t, \\
    \ell_N(x_N) &=
    \frac{1}{2}x_N^\top Q_Nx_N + q_N^\top x_N
\end{align}
\end{subequations}
for a discrete dynamics along the time horizon $t=0,\ldots,N$, with state vector $x_t \in \RR^{n_x}$, control input $u_t \in \RR^{n_u}$. The parameters of the problem are:
\begin{itemize}
	\renewcommand\labelitemi{--}
	\item dynamics matrices $A_t, E_t \in \RR^{n_x\times n_x}$, $B_t \in \RR^{n_x\times n_u}$, and $f_t\in\RR^{n_x}$,
	\item constraint matrices $C_t \in \RR^{n_c^t\times n_x}$, $D_t \in \RR^{n_c^t\times n_u}$, $h_t \in \RR^{n_c^t}$ ($n_c^t\in\NN$ being the number of rows),
	\item cost matrices $Q_t \in \symmat{n_x}$, $R_t \in \symmat{n_u} $, $S_t \in \RR^{n_x\times n_u}$, $q_t\in\RR^{n_x}$ and $r_t\in\RR^{n_u}$, and
	\item initial constraint matrix \mbox{$G_0 \in \RR^{n_g\times n_x}$} and vector $g_0 \in \RR^{n_g}$.
\end{itemize}
In most cases, the initial constraint would be $x_0 - \bar{x} = 0$ ($G_0 = -I$, $g_0 = \bar{x} \in \RR^{n_x}$).

\begin{remark}
We consider the general case of implicit dynamics~\eqref{eq:lqr:dyn}, with the particular case of explicit dynamics obtained by $E_t=-I$.
\end{remark}
\begin{remark}
	The LQ problem~\eqref{eq:lqr:problem} is not always feasible.
\end{remark}

\subsection{Lagrangian and KKT conditions}

We introduce the following Hamiltonian function: 
\begin{equation}\label{eq:lqr:hamiltonian}
\begin{split}
    H_t(x, u, \nu, \lambda) &\defeq \ell_t(x, u) + \lambda^\top (A_tx + B_tu + f_t) \\
    &\ + \nu^\top (C_tx + D_tu + h_t)
\end{split}
\end{equation}
for $t=0,\ldots,N-1$, and terminal stage Lagrangian
\begin{equation*}
	\scrL_N(x, \nu) \defeq \ell_N(x) + \nu^\top (C_Nx + h_N)
\end{equation*}
Using this notation, the Lagrangian of Problem~\eqref{eq:lqr:problem} reads
\begin{equation}\label{eq:lqr:lagrangian}
\begin{split}
    &\scrL(\bmx, \bmu, \bmnu, \bmlam) \defeq \lambda_0^\top (G_0x_0 + g_0) \\
   	&\quad+ \sum_{t=0}^{N-1} H_t(x_t, u_t, \nu_t, \lambda_{t+1}) + \lambda_{t+1}^\top E_tx_{t+1} + \scrL_N(x_N, \nu_N).
\end{split}
\end{equation}

Problem~\eqref{eq:lqr:problem} has linear constraints, hence linear constraint qualifications apply and the Karush-Kuhn-Tucker (KKT) optimality conditions read (with, for convenience, $E_{-1} = G_0$):
\begin{subequations}\label{eq:lqr:kkt}
\begin{align}
    \label{eq:lqr:kkt:adj}
    -E_{t-1}^\top\lambda_t &= Q_tx_t + S_t u_t + A_t^\top\lambda_{t+1} + C_t^\top\nu_t + q_t \\
    \label{eq:lqr:kkt:uopt}
    0 &= S_t^\top x_t + R_tu_t + B_t^\top\lambda_{t+1} + D_t^\top\nu_t + r_t \\
    \label{eq:lqr:kkt:cstr}
    0 &= C_tx_t + D_tu_t + h_t \\
    0 &= A_tx_t + B_tu_t + E_tx_{t+1} + f_t
\end{align}
for $0\leq t \leq N-1$ with terminal and initial conditions:
\begin{align}
    -E_{N-1}^\top\lambda_N &= Q_Nx_N + C_N^\top\nu_N + q_N \\
    0 &= C_Nx_N + h_N \\
    0 &= G_0x_0 + g_0.
\end{align}
\end{subequations}

\subsection{Proximal regularization of the LQ problem}

In this subsection, we introduce a proximal regularization of the LQ problem \eqref{eq:lqr:problem} in its dual variables and derive the corresponding optimality conditions.
Leveraging proximal regularization in constrained optimization settings is a generic way to tackle ill-posed problems (e.g., rank deficient constraints) in the optimization literature~\cite{nocedalNumericalOptimization2006,boydProximalAlgorithms}.
The saddle-point formulation of \eqref{eq:lqr:problem} is
\begin{equation*}
	\min_{\bmx,\bmu} \max_{\bmlam,\bmnu} \scrL(\bmx, \bmu, \bmnu, \bmlam).
\end{equation*}
The corresponding dual proximal-point iteration from previous estimates $(\bmlam^e, \bmnu^e)$ of the co-state and path multipliers is:
\begin{equation}\label{eq:lqr:dual_rprox}
    \min_{\bmx,\bmu} \max_{\bm\lambda, \bm\nu}
    \scrL(\bmx, \bmu, \bmnu, \bmlam) - \frac{\mu}{2}\left \| [\bmlam, \bmnu] - [\bmlam^e, \bmnu^e] \right \|_2^2,
\end{equation}
where $\mu > 0$ is the proximal parameter.
This \emph{proximal LQ problem} is known~\cite{rockafellarAugmentedLagrangiansApplications1976} to be equivalent to finding a minimizer of the augmented Lagrangian associated with \eqref{eq:lqr:problem}.
We denote by $\bar{g}_0$, $\bar{f}_t$ and $\bar{h}_t$ the ``shifted" right-hand side quantities
\begin{equation}
	\bar{g}_0 \defeq g_0 + \mu\lambda^e_0,\
	\bar{f}_t \defeq f_t + \mu\lambda^e_{t+1},\
	\bar{h}_t \defeq h_t + \mu\nu^e_t.
\end{equation}
Then, the KKT conditions of the proximal LQ problem are given by \eqref{eq:lqr:kkt:adj}, \eqref{eq:lqr:kkt:uopt} and
\begin{subequations}\label{eq:lqr:kkt:reg}
\begin{align}
	0 &= A_tx_t + B_tu_t + E_tx_{t+1} + \bar{f}_t - \mu\lambda_{t+1} \\
	0 &= C_tx_t + D_tu_t + \bar{h}_t - \mu\nu_t \\
	0 &= C_Nx_N + \bar{h}_N - \mu\nu_N \\
	0 &= G_0x_0 + \bar{g}_0 - \mu\lambda_0,
\end{align}
\end{subequations}
This system of equations arises when trying to solve \eqref{eq:lqr:problem} through a proximal iteration scheme, but it also arises when applying an augmented Lagrangian method to a general nonlinear control problem as in~\cref{sec:nonlinear}.

A Riccati-like recursion has been proposed in \cite{jalletImplicitDifferentialDynamic2022,jalletConstrainedDifferentialDynamic2022} to solve the proximal LQ problem. Yet, it requires solving at each stage of the recursion a larger linear system than the classical (unconstrained, unregularized) Riccati setting.

\section{Riccati Equations for the proximal LQ Problem}%
\label{sec:riccati}

In this section, we will present a generalization of the Riccati recursion, initially introduced by~\cite{jalletImplicitDifferentialDynamic2022}, which is akin to taking successive Schur complements. This requires solving large symmetric linear systems at each stage of the recursion, for which we will provide an efficient, structure-exploiting approach in~\cref{sec:blocksparse}.
A self-contained refresher of the classical Riccati recursion is given in \cref{appx:base_riccati}.

\subsection{Solving the proximal LQ by block substitution}
\label{sec:riccati:derivation}

\tikzset{highlight/.style={rectangle,
		fill=red!15,
		rounded corners = 0.5 mm,
		inner sep=1pt,
		fit=#1}}

The KKT conditions can be rewritten as a block-banded linear system of the form:
\begin{equation}\label{eq:riccati:matrix}
    \resizebox{0.915\linewidth}{!}{$
    \begin{bNiceArray}[first-row]{ccc c cccc c ccc|c}
    	\RowStyle{\arrstyle}
    	x_0 & u_0 & \nu_0 & \lambda_1 &
    	x_1 & u_1 & \nu_1 & \lambda_2 & \ldots & \lambda_N & x_N & \nu_N \\
    	Q_0     & S_0   & C_0^\top&  A_0^\top   &  &&&& &&&& q_0\\
    	S_0^\top& R_0   & D_0^\top&  B_0^\top  &               &&&&& &&& r_0\\
    	C_0 & D_0 & -\mu I	&&&&&&&&&&  \bar{h}_0 \\
    	A_0 & B_0  &   & -\mu I   & E_0      & &&&&&&& \bar{f}_0 \\
    	&&       & E_0^\top  & Q_1 & S_1&C_{1}^\top& A_1^\top &&&&& q_1\\
    	&&       &           & S_1^\top & R_1 & D_{1}^\top& B_1^\top &&&&&r_1\\
    	&&       &           & C_{1}    & D_{1} & -\mu I  &&&&&& \bar{h}_1    \\
    	&&       &           & A_1		& B_1 & &-\mu I & E_1 & \ddots &&& \bar{f}_1\\
    	&&		&	&	&	&& E_1^\top & \ddots & & & & \vdots \\
    	&&			&	&		&	&	& & \ddots & & E_{N-1} & & \bar{f}_{N-1} \\
    	&	&       &           &               &&& & & E_{N-1}^\top  & Q_N & C_N^\top &  q_N \\
    	&	&       &           &               &&& &    &  & C_N &-\mu I & \bar{h}_N
    \end{bNiceArray}
	$}
\end{equation}
For simplicity of presentation, we will consider, in this section, the case where $N = 2$.

\subsubsection{Terminal stage}
Starting from the lower-right block in the unknowns $(x_2, \nu_2)$, we can express the terminal system in the unknowns $(x_2, \nu_2)$,
\begin{equation}\label{eq:riccati:terminal}
	\begin{bNiceArray}[first-row]{cc|l}
		\RowStyle{\arrstyle}
		x_2 & \nu_2 \\ 
		Q_2 & C_2^\top & q_2 + E_1^\top\lambda_2 \\
		C_2 & -\mu I & \bar{h}_2
	\end{bNiceArray}.
\end{equation}

A Schur complement in $x_2$ leads to $\nu_2 = \tfrac{1}{\mu}(\bar{h}_2 + C_2x_2)$ and the following equation:
\begin{equation}\label{eq:riccati:link2}
	P_2x_2 + p_2 + E_1^\top\lambda_2 = 0
\end{equation}
where
$P_2 = Q_2 + \tfrac{1}{\mu}C_2^\top C_2$ and $p_2 = q_2 + \tfrac{1}{\mu}C_2^\top\bar{h}_2$ correspond to the terminal cost-to-go matrix and vector in the classical Riccati recursion.\footnote{This can be seen as follows: if $E_1=-I$ and there are no constraints, then $P_2 = Q_2$ and $p_2=q_2$, then \eqref{eq:riccati:link2} reduces to $\lambda_2 = Q_2x_2 + q_2$ which is the terminal condition in the usual Riccati recursion, see~\cref{appx:base_riccati}.}


\subsubsection{Middle blocks}
\label{subsubsec:middleblocks}

Equations \eqref{eq:riccati:matrix} and \eqref{eq:riccati:link2} lead to the following system in $(x_1,u_1,\nu_1,\lambda_2,x_2)$ (where the unknown $\lambda_1$ is a parameter):
\begin{equation}\label{eq:riccati:big_stage_system}
    \begin{bNiceArray}[first-row]{ccccc|l}
        \RowStyle{\arrstyle}
        x_1		& u_1	& \nu_1		& \lambda_2 & x_2  \\
        Q_1     & S_1   & C_1^\top	& A_1^\top  &	& q_1 + E_0^\top\lambda_1 \\
        S_1^\top& R_1   & D_1^\top	& B_1^\top  &	& r_1 \\
        C_1		& D_1	& -\mu I	&			&	& \bar{h}_1 \\
        A_1     & B_1   &			& -\mu I    & E_1 &\bar{f}_1 \\
        &       &		& E_1^\top  & P_2 & p_2
    \end{bNiceArray}.
\end{equation}
As $x_1$ is unknown, we will solve this equation parametrically. We introduce the primal-dual feedforward (resp. feedback) gains $(k_1,\zeta_1,\omega_2,a_1)$ (resp. $(K_1,Z_1,\Omega_2,M_1)$) for the control, constraint multiplier, co-state and next state, so that the parametric solution in $(u_1,\nu_1,\lambda_2,x_2)$ is:
\begin{equation}
	\begin{aligned}
		u_1 &= k_1 + K_1x_1, \ 
		&\nu_1 = \zeta_1 + Z_1x_1 \\
		\lambda_2 &= \omega_2 + \Omega_2x_1
		\ &x_2 = a_1 + M_1x_1.
	\end{aligned}
\end{equation}
These primal-dual gains satisfy the linear system: 
\begin{equation}\label{eq:riccati:stage_kkt}
    \begin{bmatrix}
        R_1 & D_1^\top& B_1^\top  & \\
	D_1	& -\mu I  &   &   \\
	B_1 &		& -\mu I    & E_1 \\
		&		& E_1^\top  & P_2
    \end{bmatrix}
    \begin{bmatrix}
        k_1 & K_1 \\ \zeta_1 & Z_1 \\
        \omega_2 & \Omega_2 \\
        a_1 & M_1
    \end{bmatrix} =-
    \begin{bmatrix}
        r_1 & S_1^\top \\
		\bar{h}_1 & C_1 \\
 		\bar{f}_1 & A_1 \\
		p_1 & 0
    \end{bmatrix}.
\end{equation}

\newcommand{\hatQ}{\widehat{Q}}
\newcommand{\hatS}{\widehat{S}}
\newcommand{\hatR}{\widehat{R}}
\newcommand{\hatq}{\widehat{q}}
\newcommand{\hatr}{\widehat{r}}

Once the matrix-matrix system \eqref{eq:riccati:stage_kkt} has been solved, we can substitute the expressions of $(u_1,\nu_1,\lambda_2,x_2)$ as functions of $x_1$ into the first line of \eqref{eq:riccati:big_stage_system}, which reduces to
\begin{equation}
    E_0^\top\lambda_1 + P_1x_1 + p_1 = 0
\end{equation}
where we have introduced the new cost-to-go matrix and vector for stage $t=1$:
\begin{subequations}\label{eq:riccati:cost_to_go}
\begin{align}
    P_1 &= Q_1 + S_1K_1 + C_1^\top Z_1 + A_1^\top \Omega_2 \\
    p_1 &= q_1 + S_1k_1 + C_1^\top \zeta_1 + A_1^\top \omega_2.
\end{align}
\end{subequations}
This equation has the expected form to close out the recursion. Indeed, the remaining system has the form
\begin{equation}
    \begin{bNiceArray}[first-row]{cccccc|l}
    	\RowStyle{\arrstyle}
    	\lambda_0 & x_0 & u_0 & \nu_0 & \lambda_1 & x_1  \\
    	-\mu I & G_0	&	&	&	&	& \bar{g}_0 \\
        G_0^\top & Q_0     & S_0   & C_0^\top	& A_0^\top & & q_0 \\
        & S_0^\top& R_0   & D_0^\top	& B_0^\top  &             &	r_0\\
        & C_0		& D_0	& -\mu I & &	& \bar{h}_0 \\
        & A_0     & B_0   &			& -\mu I    & E_0	&\bar{f}_0 \\
        &	&       &		& E_0^\top  & P_1	& p_1
    \end{bNiceArray}
\end{equation}
which has the same structure as the initial problem. We can iterate the previous derivation to obtain $(u_0, \nu_0, \lambda_1, x_1)$ as affine functions of $x_0$, then a cost-to-go matrix and vector $(P_0,p_0)$.

\subsubsection{Initial stage}
\label{sec:riccati:init}

If $x_0$ is fixed (e.g. in classical DDP algorithms~\cite{mayneSecondorderGradientMethod1966,tassaSynthesisStabilizationComplex2012}) then we are done.
However, if $x_0$ is not a fixed variable and is indeed a decision variable (e.g. in direct multiple-shooting schemes~\cite{diehlFastDirectMultiple2006}) along with $\lambda_0$, then they together satisfy the symmetric system:
\begin{equation}\label{eq:riccati:init}
	\begin{bNiceArray}[first-row]{cc|r}
		\RowStyle{\arrstyle}
		x_0 & \lambda_0 & \\
		P_0 & G_0^\top	& p_0 \\
		G_0 & -\mu I	& \bar{g}_0
	\end{bNiceArray}.
\end{equation}

\subsection{Riccati-like algorithm}

The overall \Cref{alg:serial_riccati} is finally obtained (for any $N\ge 2$) by repeating Step 2) (the middle blocks) of \cref{sec:riccati:derivation} from time step $t=N-1$ down to $t=0$, and handling the initial stage as previously described in Step 3). 
Although obtained differently, it is the same as proposed in~\cite{jalletConstrainedDifferentialDynamic2022}.

At each stage of the recursion, the solution of system~\eqref{eq:riccati:stage_kkt} can be obtained by dense matrix decomposition
(e.g. $LU$, $LDL^\top$, or Bunch-Kaufman~\cite{bunchStableMethodsCalculating1977}), 
which carries a time complexity of order \mbox{$\mathcal{O}((n_u + n_c + 2n_x)^3)$}.
However, if appropriate assumptions are made, a dense factorization routine can be avoided and replaced by one which exploits the block-sparsity of \eqref{eq:riccati:big_stage_system} as we will outline in the next \cref{sec:blocksparse}.

\begin{algorithm}[ht!]
	\KwData{Cost and constraint matrices $Q_t,S_t,R_t,q_t,r_t,A_t,B_t,C_t,E_t,D_t,f_t,h_t$}
	$P_N \leftarrow Q_N + \tfrac{1}{\mu}C_N^\top C_N$\;
	$p_N \leftarrow q_N + \tfrac{1}{\mu}C_N^\top\bar{h}_N$\;
	\tcp{Backward pass}
	\For{$t=N-1$ \KwTo $t=0$}{
		$[k_t,\zeta_t,\omega_{t+1},a_t, K_t,Z_t,\Omega_{t+1},M_t] \leftarrow$ solve \eqref{eq:riccati:stage_kkt}\;
            \tcp{Set cost-to-go following \eqref{eq:riccati:cost_to_go}}
            $P_t \leftarrow Q_t + S_tK_t + C_t^\top Z_t + A_t^\top \Omega_{t+1}$\;
            $p_t \leftarrow q_t + S_tk_t + C_t^\top \zeta_t + A_t^\top \omega_{t+1}$\;
	}
	$(x_0, \lambda_0) \leftarrow$ solve \eqref{eq:riccati:init} \tcp*{or impose value of $x_0$}
	
	\tcp{Forward pass}
	\For{$t=0$ \KwTo $N - 1$}{
		$u_t \leftarrow k_t + K_tx_t$\;
		$\nu_t \leftarrow \zeta_t + Z_tx_t$\;
		$\lambda_{t+1} \leftarrow \omega_{t+1} + \Omega_{t+1}x_t$ \;
		$x_{t+1} \leftarrow a_t + M_t$\;
	}
	$\nu_N = \frac{1}{\mu}(\bar{h}_N + C_Nx_N)$\;
	
	\caption{Generalized Riccati equations for proximal, constrained LQ problem}\label{alg:serial_riccati}
\end{algorithm}

\subsection{Block-sparse factorization for the stage KKT system}\label{sec:blocksparse}

Rather than applying a dense factorization procedure, we propose to leverage the block-sparse structure of \eqref{eq:riccati:big_stage_system}-\eqref{eq:riccati:stage_kkt} to build a more efficient algorithm than the one proposed in~\cite{jalletConstrainedDifferentialDynamic2022}.
It avoids solving a system of size $n_u+n_c+2n_x$ by a well-chosen substitution using the assumption that $E_t$ is invertible (with $E_t=-I$ when the dynamics are explicit).

This assumption can be motivated as follows: discretization of an ODE, as we mention in \cref{sec:nonlinear}, can lead to implicit discrete dynamics $\phi_t(x_t,u_t,x_{t+1})=0$ (this is the case with e.g. Runge-Kutta methods and variational integrators~\cite{jalletImplicitDifferentialDynamic2022}). Provided a small enough timestep, the solution to this implicit equation is unique: in fact there is\footnote{This result stems from the Implicit Function Theorem.} a differentiable, local inverse map $F$ defined on a neighbourhood $\calO$ of $(x_t,u_t)$ such that $\phi_t(x,u,F(x,u)) = 0$ for $(x,u)\in\calO$.

A possible alternative would be to assume that the ``cost-to-go" matrix $P_2$ in \eqref{eq:riccati:big_stage_system} is nonsingular, and compute its inverse to perform a Schur complement. The assumption can be satisfied by requiring that the pure-state cost matrices $Q_t$ be definite positive. This is a drawback for problems with e.g. semidefinite terminal cost Hessians (for instance, when only the joint velocities are penalized but not the joint configurations).

The mathematical derivation for the block-sparse factorization is postponed to \cref{app:blocksparse} as its understanding is not necessary to continue on to the next section. 
We present this as a secondary contribution of this paper, which we have implemented and evaluated in the experimental section.

\subsection*{Discussion}

In conclusion, this section reformulated the backward recursion for the proximal LQ problem introduced in~\cite{jalletConstrainedDifferentialDynamic2022} and allowed us to suggest a more efficient alternative.
More importantly, this formulation paves the road towards our parallel algorithm.
Next section will develop a variant of~\cref{alg:blocksparse_riccati} geared toward parametric problems, which will be the cornerstone to design our parallel formulation.

\section{Extension to Parametric LQ problems}%
\label{sec:parametric}

In this section, we consider problems for which the Lagrangian has an additional affine term with parameter vector $\theta \in \RR^{n_{\theta}}$.
Differentiable, parametric optimal control in general has been covered in recent literature~\cite{amosDifferentiableMPCEndtoend,dantecFirstOrderApproximation2022,oshinParameterizedDifferentialDynamic2022,bounouLeveragingProximalOptimization2023}.
In particular, the case of constrained problems with proximal regularization has been covered by Bounou et al.~\cite{bounouLeveragingProximalOptimization2023}, where~\eqref{eq:riccati:big_stage_system}.
In this subsection, we extend the block-sparse approach we presented in~\cref{sec:blocksparse} to parametric problems. We will build upon this in the next section to derive a method for parallel resolution of LQ problems~\eqref{eq:lqr:problem}.

\subsection{Parametric Lagrangians}

\def\bmxi{\bm{\xi}}

In this section, we consider a parametric LQ problem with a Lagrangian of the form (without initial condition):
\begin{equation}\label{eq:param:lagrangian}
\begin{split}
    \scrL( &\bmx, \bmu, \bmnu, \bmlam; \theta) = \bar{\scrL}_N(x_N,\nu_N;\theta) \\
    &\ + \sum_{t=0}^{N-1} \bar{H}_t(x_t, u_t, \nu_t, \lambda_{t+1}; \theta) + \lambda_{t+1}^\top E_tx_{t+1}
\end{split}
\end{equation}
where each parametric Hamiltonian $\bar{H}_t$ reads
\begin{multline}
    \bar{H}_t(x, u, \nu, \lambda; \theta) = \\
    H_t(x, u, \nu, \lambda) + \theta^\top(\Phi_{t}^\top x + \Psi_{t}^\top u + \gamma_t)
    + \tfrac{1}{2}\theta^\top \Gamma_t\theta
\end{multline}
where $H_t$ contains the non-parametric terms of the Hamiltonian, $\gamma_t \in \RR^{n_{\theta}}$ and $\Phi_{t} \in \RR^{n_x\times n_\theta}$, $\Psi_{t} \in \RR^{n_u\times n_\theta}$, \mbox{$\Gamma_t \in \RR^{n_\theta\times n_\theta}$}. Similarly, the parametric terminal Lagrangian reads
\begin{equation}
    \bar{\scrL}_N(x, \nu; \theta) = \scrL_N(x, \nu) + \theta^\top (\Phi_{N}^\top x + \gamma_N) + \frac{1}{2}\theta^\top \Gamma_N\theta.
\end{equation}
This representation can stem from having additional affine cost terms in the LQ problem. We could seek to compute the sensitivities of either the optimal value or the optimal primal-dual trajectory $\bmxi^\star = (\bmx^\star, \bmu^\star, \bmnu^\star, \bmlam^\star)$ with respect to these additional terms~\cite{amosDifferentiableMPCEndtoend}.

We denote by $\calE$ the value function of problem~\eqref{eq:lqr:problem} under parameters $\theta$, defined in saddle-point form by
\begin{equation}
    \calE(x_0, \theta) \defeq
    \min_{\bmx, \bmu}\max_{\bmnu,\bmlam}
    \scrL(\bmx, \bmu, \bmnu, \bmlam; \theta)
\end{equation}
Similarly, we denote by $\calE_\mu(x_0, \theta)$ the value function of the dual proximal-regularization~\eqref{eq:lqr:dual_rprox} (as in \cref{sec:elq_problem}) for $\mu > 0$:
\begin{equation}\label{eq:param:optimalvalue}
    \calE_\mu(x_0,\theta) \defeq 
    \min_{\bmx, \bmu}\max_{\bmnu,\bmlam}
    \scrL(\bmx, \bmu, \bmnu, \bmlam; \theta)
    - \frac{\mu}{2}\| [\bmlam,\bmnu] - [\bmlam_e,\bmnu_e] \|^2.
\end{equation}

\subsection{Expression of the value function $\calE_\mu(x_0,\theta)$}

Denote $\bmxi = (\bmx, \bmu, \bmnu, \bmlam)$ the collection of primal-dual variables.
We have the following property, derived from the Schur complement lemma (see \cref{appx:schur}):
\begin{prop}\label{prop:param_value}
    $\calE_\mu$ is a quadratic function in $(x_0, \theta)$.
    There exist $\sigma_0\in\RR^{n_{\theta}}$, $\Lambda_0\in\RR^{n_x\times n_\theta}$ and $\Sigma_0 \in \RR^{n_\theta\times n_\theta}$ such that
    \begin{equation}\label{eq:param:value_quadratic}
        \calE_\mu(x_0, \theta) = \frac{1}{2}\begin{bmatrix}
            x_0 \\ \theta
        \end{bmatrix}^\top
        \begin{bmatrix}
            P_0 & \Lambda_0 \\
            \Lambda_0^\top & \Sigma_0
        \end{bmatrix}
        \begin{bmatrix}
            x_0 \\ \theta
        \end{bmatrix}
        + p_0^\top x_0 + \sigma_0^\top \theta.
    \end{equation}
    \begin{proof}
    	The min-maximand in \eqref{eq:param:optimalvalue} is a quadratic function in $(\bmxi, \theta)$ which is convex-concave in $\bmxi$.
    	Then, we apply the lemma in \cref{appx:schur:2} to \eqref{eq:param:optimalvalue}, choosing $(x_0, \theta)$ as the parameter ($z$ in the lemma) in the saddle-point.
    \end{proof}
\end{prop}

Furthermore, the lemma in~\cref{appx:schur:2} yields
\begin{equation}\label{eq:param:schur_mat}
\begin{aligned}
    \sigma_0 &= \scrL_{\theta} - \scrL_{\theta\bmxi}\scrL_{\bmxi\bmxi}^{-1}\scrL_{\bmxi\theta} \\
    \Lambda_0 &= \scrL_{x_0\theta} - \scrL_{x_0\bmxi}\scrL_{\bmxi\bmxi}^{-1}\scrL_{\bmxi\theta} \\
    \Sigma_0 &= \scrL_{\theta\theta} - \scrL_{\theta\bmxi}\scrL_{\bmxi\bmxi}^{-1}\scrL_{\bmxi\theta},
\end{aligned}
\end{equation}
where the following derivatives in $ \theta$ are simple:
\begin{equation}
    \scrL_\theta = \sum_{t=0}^N \gamma_t,
    \ 
    \scrL_{\theta\theta} = \sum_{t=0}^N \Gamma_t, \
    \scrL_{x_0\theta} = \Phi_0.
\end{equation}
Solving $\bmxi^0 = -\scrL_{\bmxi\bmxi}^{-1}\scrL_{\bmxi}$ is handled by the variant of the Riccati recursion introduced in the previous \cref{sec:riccati}.

Denote by $\partial_\theta\bmxi=(\partial_\theta\bmx,\partial_\theta\bmu,\partial_\theta\bmnu,\partial_\theta\bmlam) = -\scrL_{\bmxi\bmxi}^{-1}\scrL_{\bmxi\theta}$ the sensitivity matrix, such that the solution of \eqref{eq:param:optimalvalue} is
\[
    \bmxi^\star(\theta) = \bmxi^0 + \partial_\theta\bmxi \cdot\theta.
\]
Solving for $\partial_\theta\bmxi $ requires a slight alteration of our generalized Riccati recursion.

The terminal stage is
\begin{equation}
    \begin{bNiceArray}[first-row]{cc|l}
        \RowStyle{\arrstyle}
        \partial_\theta x_N & \partial_\theta \nu_N & \\
        Q_N & C_N^\top & \Phi_{N} + E_{N-1}^\top \partial_\theta\lambda_N \\
        C_N & -\mu I & 0
    \end{bNiceArray}.
\end{equation}
Defining $\Lambda_N \defeq \Phi_{N} \in \RR^{n_x\times n_\theta}$, this leads to
\[
    P_N \partial_\theta x_N + \Lambda_N + E_{N-1}^\top \partial_\theta\lambda_N = 0.
\]

For non-terminal stages, we obtain a system of equations
\begin{equation}
    \begin{bNiceArray}[first-row]{ccccc|l}
    \RowStyle{\arrstyle}
    \partial_\theta x_t	& \partial_\theta u_t& \partial_\theta\nu_t & \partial_\theta\lambda_{t+1} & \partial_\theta x_{t+1}  \\
    Q_t     & S_t   & C_t^\top	& A_t^\top  &	& \Phi_{t} + E_{t-1}^\top \partial_\theta \lambda_t \\
    S_t^\top& R_t   & D_t^\top	& B_t^\top  &	& \Psi_{t} \\
    C_t		& D_t	& -\mu I	&			&	& 0 \\
    A_t     & B_t   &			& -\mu I    & E_t & 0 \\
    &       &		& E_t^\top  & P_{t+1} & \Lambda_{t+1}
    \end{bNiceArray}
\end{equation}
which can be solved as before, by introducing a matrix right-hand side variant of \eqref{eq:riccati:stage_kkt}:
\begin{equation}\label{eq:param:stage_kkt_param}
    \begin{bNiceArray}[first-row]{cccc|l}
    \RowStyle{\arrstyle}
    K_t^\theta& Z_t^\theta& \Omega^\theta_{t+1} & M_{t+1}^\theta  \\
    R_t   & D_t^\top	& B_t^\top  &	& \Psi_{t} \\
    D_t	& -\mu I	&			&	& 0 \\
    B_t   &			& -\mu I    & E_t & 0 \\
    &		& E_t^\top  & P_{t+1} & \Lambda_{t+1}
    \end{bNiceArray}
\end{equation}
such that
\begin{equation}\label{eq:param:forward_sens}
    \begin{bmatrix}
        \partial_\theta u_t \\
        \partial_\theta \nu_t \\
        \partial_\theta \lambda_{t+1} \\
        \partial_\theta x_{t+1}
    \end{bmatrix}=
    \begin{bmatrix}
        K_t\partial_\theta x_t + K_t^\theta \\
        Z_t\partial_\theta x_t + Z_t^\theta \\
        \Omega_{t+1}\partial_\theta x_t + \Omega_{t+1}^\theta \\
        M_t\partial_\theta x_t + M_t^\theta
    \end{bmatrix}.
\end{equation}
This leads to a reduced first line,
\begin{multline}
    \underbrace{
        (Q_t + S_tK_t + C_t^\top Z_t)
    }_{=P_t}\partial_\theta x_t +  \\
    \underbrace{%
        \Phi_{t} + S_tK_t^\theta + C_t^\top Z_t^\theta%
    }_{\defeq \Lambda_t} + E_{t-1}^\top \partial_\theta\lambda_t = 0
\end{multline}
where $\Lambda_t \in \RR^{n_x\times n_\theta}$, thereby closing the recursion.

The recursion continues until reaching the initial stage, which gives the expression corresponding to $\nabla_{x_0}\calE(x_0,\theta)$:
\begin{equation*}
    P_0x_0 + \Lambda_0\theta + p_0.
\end{equation*}
Then, the sensitivity matrix $\partial_\theta\bmxi$ can be extracted by a forward pass, iterating \eqref{eq:param:forward_sens} for $t=0,\ldots,N-1$.

Furthermore, the remaining value function parameters $(\sigma_0,\Sigma_0)$ can be obtained by backward recursion:
\begin{prop}
    The vector $\sigma_0$ and matrix $\Sigma_0$ from \eqref{eq:param:value_quadratic} can be computed as follows: let
    \begin{subequations}
    \begin{align}
        \sigma_N &= \gamma_N \\
        \Sigma_N &= \Gamma_N
    \end{align}
    and for $t=N-1,\ldots,0$,
    \begin{align}
        \sigma_t &= \sigma_{t+1} + \gamma_t + \Psi_{t}^\top k_t + \Lambda_{t+1}^\top a_t \\
        \Sigma_t &= \Sigma_{t+1} + \Gamma_t + \Psi_{t}^\top K_t^\theta + \Lambda_{t+1}^\top M_t^\theta.
    \end{align}
    \end{subequations}
\end{prop}

\begin{algorithm}[ht!]
    \KwData{Cost and constraint matrices and vectors $Q_t, S_t, R_t, q_t, r_t, A_t, B_t, C_t, D_t, f_t, h_t$, parameters $\Phi_{N},\gamma_N$}
    $P_N \leftarrow Q_N + \tfrac{1}{\mu}C_N^\top C_N$\;
    $p_N \leftarrow q_N + \tfrac{1}{\mu}C_N^\top\bar{h}_N$\;
        {\color{MidnightBlue} $\Sigma_N \leftarrow \Gamma_N$\;}
    {\color{MidnightBlue} $\Lambda_N \leftarrow \Phi_{N}$\;}
    {\color{MidnightBlue} $\sigma_N \leftarrow \gamma_N$\;}
    \tcp{Backward pass}
    \For{$t=N-1$ \KwTo $t=0$}{
        $[K_t^\theta, Z_t^\theta, \Omega_{t+1}^\theta, M_t^\theta] \leftarrow$ solve \eqref{eq:param:stage_kkt_param}\;
        $P_t \leftarrow Q_t + S_tK_t + C_t^\top Z_t + A_t^\top \Omega_{t+1}$\;
        $p_t \leftarrow q_t + S_tk_t + C_t^\top \zeta_t + A_t^\top\omega_{t+1}$\;
        {\color{MidnightBlue}$\Sigma_t \leftarrow \Gamma_t + \Sigma_{t+1} + \Psi_{t}^\top K_t^\theta + \Lambda_{t+1}^\top M_t^\theta$}\;
        {\color{MidnightBlue}$\Lambda_t \leftarrow \Phi_{t} + K_t^\top \Psi_{t} + M_t^\top \Lambda_{t+1}$}\;
        {\color{MidnightBlue}$\sigma_t \leftarrow \sigma_{t+1} + \gamma_t + \Psi_{t}^\top k_t + \Lambda_{t+1}^\top a_t$}\;
    }
    \SetKwFunction{ComputeInitial}{ComputeInitial}
    $(x_0, \lambda_0, \theta) \leftarrow \ComputeInitial()$\;
	
    \tcp{Forward pass}
    \For{$t=0$ \KwTo $N - 1$}{
        $u_t \leftarrow k_t + K_tx_t + {\color{MidnightBlue} K^\theta_t\theta}$\;
        $\nu_t \leftarrow \zeta_t + Z_tx_t + {\color{MidnightBlue} Z^\theta_t\theta}$\;
        $\lambda_{t+1} \leftarrow \omega_{t+1} + \Omega_{t+1} x_t + {\color{MidnightBlue}\Omega_{t+1}^\theta \theta}$\;
        $x_{t+1} \leftarrow a_t + M_tx_t + {\color{MidnightBlue} M_t^\theta \theta}$\;
    }
    $\nu_N = \frac{1}{\mu}(\bar{h}_N + C_Nx_N)$\;
	
    \caption{Generalized Riccati recursion for parametric problems}\label{alg:param_riccati}
\end{algorithm}

\paragraph*{Algorithm}
Bringing this all together, we outline a parametric generalized Riccati recursion in \cref{alg:param_riccati}, which provides the solution of \eqref{eq:param:optimalvalue}.
Step 12 (\ComputeInitial) of \cref{alg:param_riccati} is the procedure where the choice of parameter $\theta \in \RR^{n_\theta}$, initial state $x_0$, and initial co-state $\lambda_0$ is made: $\theta$ and $x_0$ could be some fixed values, or all three could be decided jointly by satisfying an equation. This choice is application-dependent.

\section{Parallelization of the LQ solver}%
\label{sec:parallel}

\def\maxproc{J}
\def\xbar{\tilde{x}}
\def\lbar{\tilde{\lambda}}
\def\Lbar{\tilde{\Lambda}}
\def\Pbar{\tilde{P}}
\def\pbar{\tilde{p}}
\def\Ebar{\tilde{E}}
\def\sgbar{\tilde{\sigma}}
\def\Sgbar{\tilde{\Sigma}}

In this section, we present a derivation for a parallelized variant of the algorithm introduced in~\cref{sec:riccati}, through the lens of solving parametric LQ problems.

We will split problem~\eqref{eq:lqr:problem} into $\maxproc + 1$ parts (or \emph{legs}) where $1 \leq \maxproc < N$.
Let $\calP = \{0 = i_0 < i_1 < \cdots < i_\maxproc < N - 1 < N\}$ be a set of partitioning indices for $\llbracket 0,N\rrbracket$. We denote $\calI_j = \llbracket i_j, i_{j+1}-1 \rrbracket$ for $j\leq\maxproc$ (where $i_{\maxproc+1} = N+1$), such that $\llbracket 0,N\rrbracket = \bigcup_{j=0}^J \calI_j$.

Each ``leg" of the split problem, except for the terminal one, will be parameterized by the co-state $\lambda_{i_j}$ (taking the place of $\theta$ in the previous \cref{sec:parametric}) which connects it to the next. In this respect, our algorithm differs from~\cite{laineParallelizingLQRComputation2019}, where the parameter is the unknown value of the first state in the next leg.

\subsection{$2$-way split ($\maxproc=1$)}

For simplicity, we start by describing how an LQ problem can be split into two parts (the case $\maxproc=1$) and explain how to merge them together.

The partition of $\llbracket 0, N\rrbracket$ is $\calI_0\cup \calI_1$ where $\calI_0 = \llbracket 0, i_1-1\rrbracket$ and $\calI_1=\llbracket i_1,N\rrbracket$, for some $1\leq i_1<N$. Denote $\lbar_1 = \lambda_{i_1}$ the co-state for the $i_1$-th dynamical constraint, which will be our splitting variable (taking the role of \cref{sec:parametric}'s parameter $\theta$), and $\xbar_1 = x_{i_1}$ the corresponding state.

We proceed by splitting up the Lagrangian of the full-horizon problem:
\begin{equation}
\begin{split}
    \scrL(\bmx, \bmu, \bmnu, \bmlam) &= \scrL^0(\bmx_{\calI_0}, \bmu_{\calI_0}, \bmnu_{\calI_0}, \bmlam_{\calI_0}; \lbar_1) +
    \lbar_1^\top E_{i_1-1} \xbar_1 \\
    &\quad +
    \scrL^1(\bmx_{\calI_1}, \bmu_{\calI_1}, \bmnu_{\calI_1}, \bmlam_{\calI_1})
\end{split}
\end{equation}
where $\scrL^0$ is the \emph{parametric} Lagrangian for the first leg (running for $t\in \calI_0$), \emph{parameterized by $\theta=\lbar_1$}:
\begin{multline}
    \scrL^0(\bmx_{\calI_0}, \bmu_{\calI_0}, \bmnu_{\calI_0}, \bmlam_{\calI_0}; \lbar_1) = \\
    \sum_{t=0}^{i_1-2} H_t(x_t, u_t, \nu_t, \lambda_{t+1}) + x_{t+1} E_t^\top \lambda_{t+1} \\
    +
    H_{i_1-1}(x_{i_1-1}, u_{i_1-1}, \nu_{i_1-1}, \lbar_1),
\end{multline}
and the non-parametric Lagrangian $\scrL^1$ for the second leg (running for $t\in\calI_1$):
\begin{multline}
    \scrL^1(\bmx_{\calI_1}, \bmu_{\calI_1}, \bmnu_{\calI_1}, \bmlam_{\calI_1}) = \\
    \sum_{t=i_1}^{N-1} H_t(x_t, u_t, \nu_t, \lambda_{t+1}) + x_{t+1} E_t^\top \lambda_{t+1}
    + \scrL_N(x_N, \nu_N).
\end{multline}

We now define their respective value functions:
\begin{subequations}
\begin{align}
    \calE^0(x_0, \lbar_1) &= \min_{\bmx,\bmu}\max_{\bmnu,\bmlam} \scrL^0(\bmx,\bmu,\bmnu,\bmlam; \lbar_1) \\
    \calE^1(\xbar_1) &= \min_{\bmx,\bmu}\max_{\bmnu,\bmlam} \scrL^1(\bmx,\bmu,\bmnu,\bmlam)
\end{align}
\end{subequations}
where the second value function does \emph{not} depend on $\lbar_1$.
Similarly to \eqref{eq:param:optimalvalue}, we can define their dual-regularized value functions $\calE_\mu^0 $ and $\calE_\mu^1$ (the former includes the $-\tfrac{\mu}{2}\|\lbar_1\|_2^2$ term).
This allows us to write the full-horizon regularized LQ problem as the following saddle-point (assuming fixed $x_0$):
\begin{equation}
    \min_{\xbar_1} \max_{\lbar_1}
    \calE_\mu^0(x_0, \lbar_1)
    + \lbar_1^\top E_{i_1-1} \xbar_1
    + \calE_\mu^1(\xbar_1).
\end{equation}

We introduce the shorthands $\Pbar_1 = P_{i_1}$, $\pbar_1 = p_{i_1}$, $\Ebar_1 = E_{i_1-1}$.
The stationarity equations for this saddle-point are
\begin{subequations}
\begin{align}
    \nabla_{\lbar_1}\calE_\mu^0(x_0, \lbar_1) + \Ebar_1\xbar_1 &= 0, \\
    \Ebar_1^\top \lbar_1 + \nabla_x\calE_\mu^1(\xbar_1)	&= 0,
\end{align}
\end{subequations}
or, substituting the expression \eqref{eq:param:value_quadratic},
\begin{equation}
    \label{eq:parallel:reduced_kkt1}
    \begin{bmatrix}
        \Sigma_0 & \Ebar_1 \\
        \Ebar_1^\top & \Pbar_1
    \end{bmatrix}
    \begin{bmatrix}
        \lbar_1 \\ \xbar_1
    \end{bmatrix} = -
    \begin{bmatrix*}[l]
        \sigma_0 + \Lambda_0^\top x_0 \\
        \pbar_1
    \end{bmatrix*}.
\end{equation}

The formulation can be further augmented if $x_0$ is a decision variable with initial constraint $G_0x_0 + g_0 = 0$ associated with multiplier $\lambda_0$: the corresponding dual-regularized system is
\begin{equation}
    \label{eq:parallel:reduced_kkt2}
	\begin{bmatrix}
		-\mu I  & G_0 & & \\
		G_0^\top& P_0 & \Lambda_0 & \\
				& \Lambda_0^\top & \Sigma_0 & \Ebar_1 \\
				& 				 & \Ebar_1^\top & \Pbar_1
	\end{bmatrix}
	\begin{bmatrix}
		\lambda_0 \\ x_0 \\ \lbar_1 \\ \xbar_1
	\end{bmatrix}
	= -\begin{bmatrix}
		\bar{g}_0 \\ p_0 \\ \sigma_0 \\ \pbar_1
	\end{bmatrix}.
\end{equation}

\paragraph*{Forward pass} Once the above linear system is solved, we can reconstruct the full solution $(\bmx, \bmu, \bmnu,\bmlam)$ by running a separate forward pass for each leg according to the equations in \cref{alg:param_riccati}.

\subsection{Generalization to $\maxproc\geq 2$}

We now consider the case $J\geq 2$.
For $0\leq j \leq \maxproc$, we denote $\xbar_j = x_{i_j}$ and $\lbar_j = \lambda_{i_j}$. Consider, for $j < \maxproc$, the value function $\calE^j(\xbar_j , \lbar_{j+1} )$ associated with the subproblem running from time $t = i_j$ to $t=i_{j+1}-1$:
\begin{equation}
    \calE^j(\xbar_j, \lbar_{j+1})
    = \min\max
    \sum_{t=i_j}^{i_{j+1}-2} H_t + \lambda_{t+1}^\top E_t x_{t+1}
    + H_{i_{j+1}-1}.
\end{equation}
For $j=\maxproc$, the value function corresponding to the subproblem running for indices $ t \in \calI_\maxproc$ is
\begin{equation}
    \calE^J(\xbar_\maxproc) = \min\max
    \sum_{t=i_\maxproc}^{N-1}
    H_t + \lambda_{t+1}^\top E_t x_{t+1}
    + \scrL_N.
\end{equation}
We also define their proximal regularizations $(\calE_\mu^j)_j$ as in \eqref{eq:param:optimalvalue}.

Finally, the dual-regularized full-horizon LQ problem is equivalent to the saddle-point in $\bm{\xbar} = (\xbar_j)_{0\leq j\leq \maxproc}$ and $\bm{\lbar}$:
\begin{equation}\label{eq:parallel:condensed_sys}
    \min_{\bm{\xbar}} \max_{\bm{\lbar}}
    \sum_{j=0}^{\maxproc-1}
    \calE_\mu^j ( \xbar_j, \lbar_{j+1} )
    + \lbar_{j+1}^\top \Ebar_{j+1} \xbar_{j+1} + \calE_\mu^J(\xbar_\maxproc).
\end{equation}
This problem also has a temporal structure, which leads to a block-tridiagonal system of equations.

We introduce the following notations, for $0\leq j\leq \maxproc$ ($j < \maxproc$ for the last four):
\begin{equation}
\begin{alignedat}{3}
    \Pbar_j &= P_{i_j} \quad
    &\pbar_j &= p_{i_j} \\
    \Ebar_{j+1} &= E_{i_{j+1}-1} \quad
    &\sgbar_j &= \sigma_{i_j} \\
    \Lbar_j &= \Lambda_{i_j} \quad
    &\Sgbar_j &= \Sigma_{i_j}
\end{alignedat}
\end{equation}
(in particular, $\pbar_0 = p_0$, $\Pbar_0=P_0$, and so on) so that, for $0 \leq j < J$,
\begin{subequations}
\begin{align}
    \nabla_{\lbar_{j+1}} \calE_\mu^j(\xbar_j, \lbar_{j+1}) &=
    \sgbar_j + \Lbar_j^\top \xbar_j + \Sgbar_j\lbar_{j+1} \\
    \nabla_{\xbar_j} \calE_\mu^j(\xbar_j, \lbar_{j+1}) &= \pbar_j + \Lbar_j\lbar_{j+1} + \Pbar_j\xbar_j
\end{align}
and
\begin{equation}
    \nabla\calE_\mu^\maxproc(\xbar_\maxproc)
    = \pbar_\maxproc + \Pbar_\maxproc\xbar_\maxproc.
\end{equation}
\end{subequations}

\begin{prop}
    The optimality conditions for \cref{eq:parallel:condensed_sys} are given by the following system of equations:
    \begin{subequations}
    \begin{align}
        \sgbar_j + \Lbar_j^\top\xbar_j + \Sgbar_j\lbar_{j+1} + \Ebar_{j+1}\xbar_{j+1} &= 0, \ 0\leq j < \maxproc \\
        \pbar_j + \Ebar_j^\top \lbar_j + \Pbar_j\xbar_j + \Lbar_j\lbar_{j+1} &= 0, \ 1\leq j < \maxproc
    \end{align}
    and
    \begin{equation}
        \Ebar_\maxproc^\top \lbar_\maxproc + \Pbar_\maxproc \xbar_\maxproc + \pbar_\maxproc = 0.
    \end{equation}
    \end{subequations}

\end{prop}

The system of equations can be summarized as the following sparse system:
\begin{equation}\label{eq:parallel:consensus_tridiag}
    \begin{bNiceArray}[first-row]{ccccccc|l}
        \RowStyle{\arrstyle}
        \lambda_0 & x_0 & \lbar_1 & \xbar_1 & \lbar_2 & \cdots & \xbar_\maxproc & \\
        -\mu I & G_0 & & & & & & \bar{g}_0 \\
        G_0^\top & P_0 & \Lambda_0 & & & & & p_0 \\
        & \Lambda_0^\top & \Sigma_0 & \Ebar_1 & & & & \sigma_0 \\
        & & \Ebar_1^\top & \Pbar_1 & \Lbar_1 & & & \pbar_1  \\
        & & & \Lbar_1^\top & \ddots & & & \sgbar_1 \\
        & & & & & \ddots & \Ebar_\maxproc & \vdots \\
        & & & & & \Ebar_\maxproc^\top & \Pbar_\maxproc & \pbar_\maxproc
    \end{bNiceArray}
\end{equation}

This system above can be solved by sparse $LDL^\top$ factorization, or by leveraging a specific block-tridiagonal algorithm.
We will denote $\mathbb{M}$ the matrix in \eqref{eq:parallel:consensus_tridiag}.

The overall method is summarized in~\cref{alg:parallel_lqr}.


\begin{algorithm}[ht!]
    \KwIn{}
    \For{$j=0,\ldots,J$ in parallel}{%
        \tcp{Parameterize last stage of each leg}
        $F_{x,i_j} \leftarrow A_{i_j}^\top$\;
        $F_{u,i_j} \leftarrow B_{i_j}^\top$\;
        $\gamma_{i_j} \leftarrow \bar{f}_{i_j}$\;
        \tcp{Solve \eqref{eq:parallel:condensed_sys} parametrically in $\lbar_j$}
        Compute $(\Pbar_j, \Lbar_j, \Sgbar_j, \pbar_j, \sgbar_j)$ using \cref{alg:param_riccati}\;
    }
    
    $(\xbar_j)_j, (\lbar_j)_j \leftarrow$ solve consensus system \eqref{eq:parallel:consensus_tridiag}\;
    \For{$0=1,\ldots,J$ in parallel}{%
        Compute the forward pass of \eqref{eq:parallel:condensed_sys} starting from $\xbar_j$\; 
    }
    \caption{Parallel condensation LQR}\label{alg:parallel_lqr}
\end{algorithm}

\subsection{Block-tridiagonal algorithm for the reduced system}

In this subsection, we present an efficient algorithm for solving the reduced linear system, exploiting its symmetric, block-tridiagonal structure by employing a block variant of the Thomas algorithm.
The diagonal is $(-\mu I, \Pbar_0, \Sgbar_0, \ldots, \Pbar_J)$, and the superdiagonal is $(G_0, \Lbar_0, \Ebar_1, \ldots, \Ebar_J)$. The elimination order will be from back-to-front, constructing a block-sparse $UDU^\top$ decomposition of the matrix $\mathbb{M}$. The resulting routine has linear complexity $\mathcal{O}(J)$ in the number of processors.

In a recent related paper, \citet{jordanaStagewiseImplementationsSequential2023} leverage the same algorithm to restate the classical Riccati recursion.

\section{Implementation in a nonlinear trajectory optimizer}
\label{sec:nonlinear}

We now consider a nonlinear discrete-time trajectory optimization problem with implicit system dynamics:
\begin{subequations}\label{eq:nocp}
	\begin{align}
		\min_{\bmx,\bmu} &~J(\bmx, \bmu) =
		\sum_{t=0}^{N-1} \ell_t(x_t, u_t) + \ell_N(x_N) \\
		\subjto & x_0 = x^0 \\
		& \phi_t(x_t, u_t, x_{t+1}) = 0 \\
		& h_t(x_t, u_t) \leq 0 \\
		& h_N(x_T) \leq 0.
	\end{align}
\end{subequations}
The implicit discrete dynamics $\phi_t(x_t, u_t, x_{t+1}) = 0$ is often a discretization scheme for an ODE $\dot{x} = f(t, x, u)$ or implicit ODE $f(t,x,u,\dot{x}) = 0$ (e.g. implicit Runge-Kutta methods).

Following~\cite{jalletConstrainedDifferentialDynamic2022}, we use a proximal augmented Lagrangian scheme to solve this problem.
This solver implements an outer (proximal, augmented Lagrangian) loop which iteratively solves a family of proximal LQ problems \eqref{eq:lqr:dual_rprox} \emph{instead} of the initial LQ problem \eqref{eq:lqr:problem}. The coefficients of the problem are obtained from the derivatives of \eqref{eq:nocp} with the following equivalences: 
%
%
\begin{subequations}
\begin{align*}
    A_t &= \phi_{x,t} \
    B_t = \phi_{u,t} \
    E_t = \phi_{y,t} \\
    C_t &= h_{x,t} \
    D_t = h_{u,t} \\
 f_t &= \phi_t(x_t,u_t,x_{t+1}) \\
	q_t &= \ell_{x,t} + A_t^\top \lambda_{t+1} + C_t^\top\nu_t + E_{t-1}^\top \lambda_t \\
	r_t &= \ell_{u,t} + B_t^\top \lambda_{t+1} + D_t^\top\nu_t \\
	Q_t &= \ell_{xx,t} + \lambda_{t+1} \cdot \phi_{xx,t} + \nu_t\cdot h_{xx,t} + \lambda_t \cdot \phi_{yy,t-1} \\
	R_t &= \ell_{uu,t} + \lambda_{t+1} \cdot \phi_{uu,t} + \nu_t\cdot h_{uu,t} \\
	S_t &= \ell_{xu,t} + \lambda_{t+1} \cdot \phi_{xu,t} + \nu_t\cdot h_{xu,t},
\end{align*}
\end{subequations}
All these quantities are directly obtained from applying a semi-smooth primal-dual Newton step, as detailed in~\cite{jalletPROXDDPProximalConstrained2023}\footnote{We have neglected two second-order terms corresponding to the derivatives $\partial^2_{u_tx_{t+1}}\scrL = \lambda_{t+1} \cdot \phi_{uy,t}(x_t, u_t,x_{t+1}) $ and $\partial^2_{x_tx_{t+1}}\scrL = \lambda_{t+1}\cdot \phi_{xy,t}(x_t,u_t,x_{t+1})$. They could be added to the block-sparse factorization of~\cref{sec:blocksparse} with further derivations which are outside the scope of this paper.}

We then implemented our parallel algorithm for solving the proximal LQ.
Our implementation will be open-sourced upon acceptance of the paper.
It is then straightforward to adapt an implementation of \cite{jalletImplicitDifferentialDynamic2022} by replacing the proximal LQ solver by the parallel formulation. This also makes it possible to fairly compare the new algorithm with the original formulation.

\section{Discussion}

In the introduction, we mentioned a few prior methods for parallelizing the resolution of system \eqref{eq:lqr:kkt}, and made a distinction between indirect and direct methods -- the terms ``indirect" and ``direct" are used in the linear algebra sense.
In this section, we will provide a more detailed discussion of these prior methods to distinguish with our own.

In~\cite{giftthalerFamilyIterativeGaussNewton2018} and~\cite{liUnifiedPerspectiveMultiple2023}, multiple-shooting formulations for nonlinear DTOC problems are shown to enable parallel nonlinear rollouts in the forward pass. There, multiple-shooting defects are used at specific knots of the problem (called \textit{shooting states}) to split the horizon into segments where, once the shooting state is known (through a serial linear rollout), nonlinear rollouts on each segment can be performed in parallel. Our method, however, can parallelize both the backward \emph{and} (linear) forward passes, but no strategy for a nonlinear rollout has been discussed in this paper. In light of the hybrid rollout strategy of~\cite{giftthalerFamilyIterativeGaussNewton2018}, parallel nonlinear rollouts between each ``leg" could be used in a nonlinear solver after applying our method to compute a linear policy and linear rollout in parallel.

\begin{table}[ht!]
	\centering
	\begin{tabularx}{\linewidth}{@{}l|c|X}
		\toprule
		\textbf{Method} & \textbf{Type} & Notes \\
		\midrule
		Multiple-shooting~\cite{giftthalerFamilyIterativeGaussNewton2018,liUnifiedPerspectiveMultiple2023} & Direct & Only the forward pass is parallel. \\
		\midrule[0.2pt]
		PCG~\cite{adabagMPCGPURealTimeNonlinear2023,buSymmetricStairPreconditioning2023} & Indirect & Adapted to GPUs. Exploits structure through sparse preconditioner. \\
		Gauss-Seidel~\cite{dengHighlyParallelizableNewtontype2018,plancherPerformanceAnalysisParallel2020} & Indirect & Not meant to solve~\eqref{eq:lqr:problem}, but produce iterations for~\eqref{eq:nocp}. \cite{dengHighlyParallelizableNewtontype2018} explicitly considers equality constraints. \\
		ADMM~\cite{stathopoulosHierarchicalTimesplittingApproach2013} & Indirect & ADMM splitting costs, dynamical, and state-control constraints. \\
		\midrule[0.2pt]
		State linkage~\cite{laineParallelizingLQRComputation2019} & Direct & $\bmlam$ obtained by least-squares. \\
		State-control linkage~\cite{nielsenLogParallelAlgorithm2014} & Direct & Results in subproblems similar to~\cite{wrightPartitionedDynamicProgramming1991}. \\
		Co-state split~\cite{wrightPartitionedDynamicProgramming1991,nielsenParallelStructureExploiting2015} & Direct & \cite{nielsenParallelStructureExploiting2015} works on the value function parameters. \cite{wrightPartitionedDynamicProgramming1991} considers constraints (through nullspace/QR decomposition). Both condense into another instance of \eqref{eq:lqr:problem}. \\
		Co-state split (\textbf{Ours}) & Direct & Condenses into state/co-state system \eqref{eq:parallel:condensed_sys} instead of another instance of \eqref{eq:lqr:problem}. Explicitly considers equality constraints, implicit dynamics, dual proximal regularization.
	\end{tabularx}
	\caption{Comparison table of multiple methods for parallel computation of the LQ step.}
	\label{tab:parallel_methods}
\end{table}

\noindent\textbf{Indirect methods.}
As for indirect methods for solving \eqref{eq:lqr:kkt}, we distinguish a few subclasses of methods.
These methods all work towards parallelizing the entire computation of a linear search direction for \eqref{eq:nocp} by approximating a solution for \eqref{eq:lqr:problem}.
One is given by preconditioned conjugate gradient (PCG) methods such as~\cite{adabagMPCGPURealTimeNonlinear2023,buSymmetricStairPreconditioning2023} which are well-suited to GPUs, exploiting the block-banded structure to find appropriate preconditioners.
In particular~\cite{adabagMPCGPURealTimeNonlinear2023,buSymmetricStairPreconditioning2023} reduce \eqref{eq:lqr:kkt}, through a tailored Schur complement, to a system in the co-states $\bmlam$ which is solved iteratively. A major assumption there is that both cost Hessian matrices $(Q_t,R_t)$ are positive definite (there are no $(S_t)$ cross-terms, and equality constraints are not considered).
Another idea suited to nonlinear OC is to re-use a previous set of value function parameters, as proposed by~\cite{plancherPerformanceAnalysisParallel2020} using the multiple-shooting forward sweep from~\cite{giftthalerFamilyIterativeGaussNewton2018}.
A similar idea is to adapt Gauss-Seidel iteration, as in~\cite{dengHighlyParallelizableNewtontype2018}, where the Riccati recursion is modified to use a previous iteration's linear relation between $\lambda_{t+1}$ and $x_t$.
Finally, \cite{stathopoulosHierarchicalTimesplittingApproach2013} propose an instance of ADMM~\cite{boydDistributedOptimizationStatistical2010} which alternatively iterates between optimizing cost and projecting onto the dynamical constraints.

\noindent\textbf{Direct methods.}
In comparison, our method is squarely in the realm of direct methods for solving \eqref{eq:lqr:kkt}. These methods exploit problem structure when applying classical factorization or (block) Gaussian elimination procedures. As presented in~\cref{sec:parallel}, our method splits the problem at specific co-states at stages $t\in\calP=(i_j)_j$, and condenses it into a saddle-point over the splitting states and co-states $(x_t,\lambda_t)_{t\in\calP}$. In contrast, \citet{wrightPartitionedDynamicProgramming1991} condenses the problem into another MPC-like subproblem which also includes controls (and is thus over $(x_t,u_t,\lambda_t)_{t\in\calP}$).
\citet{nielsenParallelStructureExploiting2015} achieve a similar set of subproblems, by partial condensing (eliminating states and keeping their unknown controls $(u_t)$) followed by reduction through SVD and parametrizing with respect to the value function parameters -- this is similar to parametrizing with respect to co-states.
\citet{laineParallelizingLQRComputation2019} condense the problem into an (overdetermined) linear system in the co-states through a linkage constraint in the states. \cite{wrightPartitionedDynamicProgramming1991,nielsenLogParallelAlgorithm2014,nielsenParallelStructureExploiting2015} show that their constructions -- which yield MPC subproblems in the same form as~\eqref{eq:lqr:problem}, can be iterated to further reduce each subproblem. In our setting, the linear subproblem~\eqref{eq:parallel:consensus_tridiag} does not have that same structure (such that our construction from~\cref{sec:parallel} cannot be iterated), however, it is still possible to leverage or design a parallel structure-exploiting routine (generic sparse e.g.,~\cite{schenkPARDISOHighperformanceSerial2001}, or specific for block-tridiagonal systems).

The different methods that have been discussed are summarized in \Cref{tab:parallel_methods}.

\section{Experiments}
\label{sec:experiments}

We have implemented the algorithms presented in this paper in \textsf{C++} using the Eigen~\cite{eigenweb} linear algebra library and the OpenMP API~\cite{chandra2001parallel} for parallel programming. This implementation has been added to our optimal control framework \textsc{aligator}\footnote{\url{https://github.com/Simple-Robotics/aligator/}}. It is the authors' aim to improve its efficiency in the future.

\subsection{Cyclic LQ problem}

\begin{figure}[ht!]
    \centering
    \includegraphics[width=.85\linewidth]{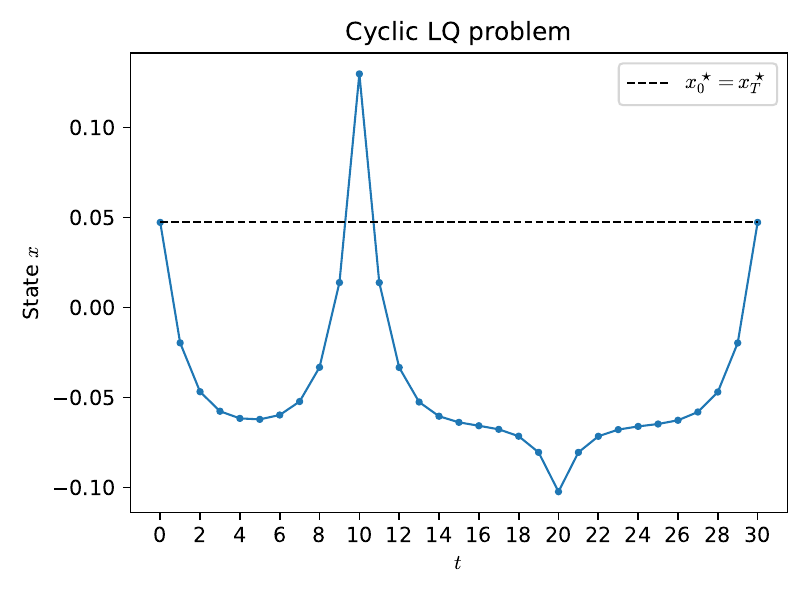}
    \caption{LQ problem with cyclical constraint $x_0=x_{30}$ ,in one dimension. No other initial condition was provided for $x_0$.}
    \label{fig:cyclic_lqr_1d}
\end{figure}

\begin{figure}[ht!]
    \centering
    \includegraphics[width=.85\linewidth]{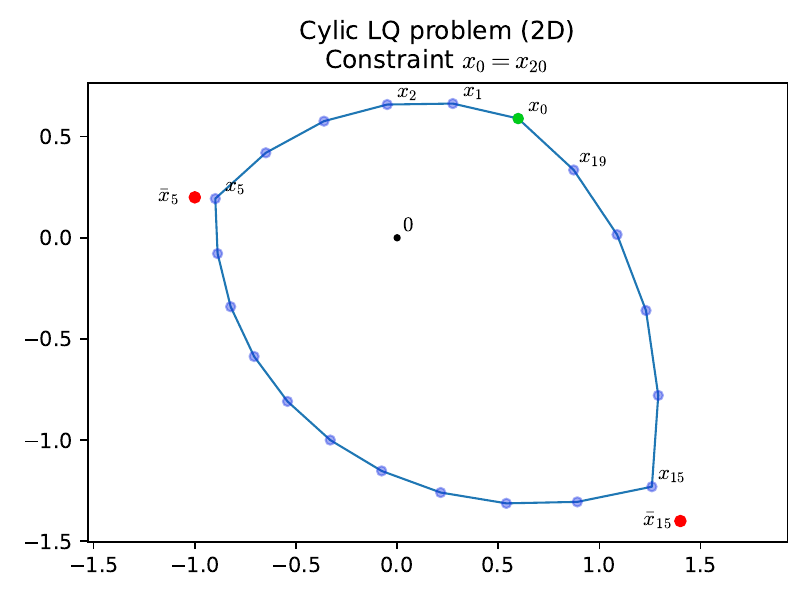}
    \caption{Cyclic LQ problem in the 2D plane. Cost function \mbox{$\ell(x,u) = 10^{-3}\|x\|^2 + \|u\|^2$ except at for $t\in\{5,15\}$} where \mbox{$\ell(x,u)=0.2\|x-\bar{x}_t\|^2 + \|u\|^2$}.}
    \label{fig:cyclic_lqr_2d}
\end{figure}

\Cref{fig:cyclic_lqr_1d,fig:cyclic_lqr_2d} show trajectories for sample cyclic LQ problems, which are formulated in \cref{sec:param:cyclic}.

\subsection{Synthetic benchmark}

\begin{figure}[ht]
    \centering
    \includegraphics[width=.96\linewidth]{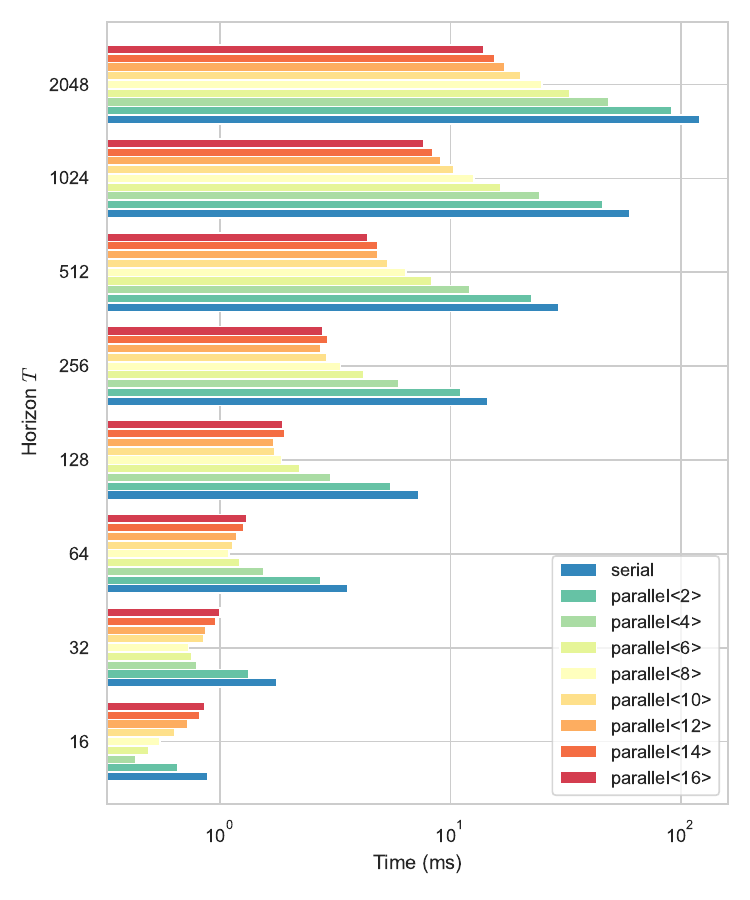}
    \caption{Timings for a backward-forward sweep of the solver on a synthetic benchmark.}
    \label{fig:exp:synth_benchmark}
\end{figure}

To assess the speedups our implementation of the parallel algorithm could achieve, we implemented a synthetic benchmark of problems with different horizons ranging from $T=16,\ldots,2048$. The benchmark was run on an Apple Mac Studio M1 Ultra, which has 20 cores (16-P cores, 4 E-cores). The resulting timings are given in \Cref{fig:exp:synth_benchmark}. These results show that our implementation is not able to reach 100\% efficiency (defined as the ratio of the speedup to the number of cores). Instead, we obtain between 50-60\% efficiency for the longer horizons, but this decreases for shorter problems. This might be due to the overhead of solving \eqref{eq:parallel:consensus_tridiag}, and that of dispatching data between cores. Greater efficiency would certainly be obtained with a more refined implementation (optimizing memory allocation, cache-friendliness).

\subsection{Nonlinear trajectory optimization}

Computation of robot dynamical quantities (joint acceleration, frame jacobians\ldots) is provided by the Pinocchio~\cite{carpentierPinocchioLibraryFast2019,carpentierProximalSparseResolution2021} rigid-body dynamics library. The \textsc{Talos} benchmark and NMPC experiments were run on a Dell \textsc{xps} laptop with an Intel i9-13900k CPU (8 P-cores and 16 E-cores).

\subsubsection{\textsc{Talos} locomotion benchmarks}

\begin{figure}[ht!]
    \centering
    \includegraphics[width=\linewidth]{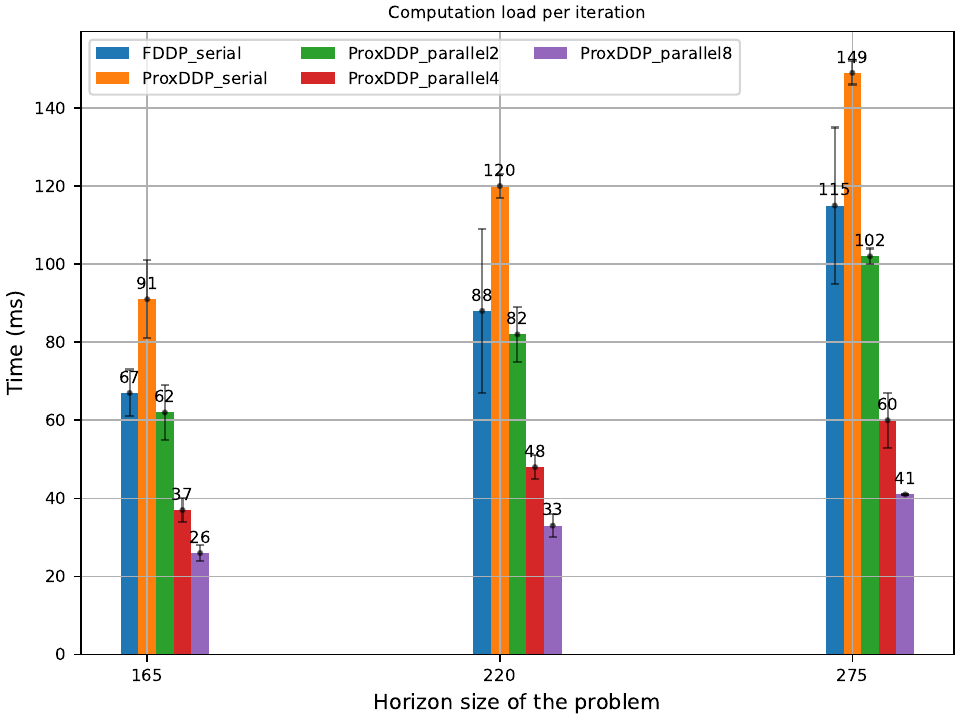}
    \caption{\textsf{C++} benchmarks of a trajectory optimization problem involving two forward steps with the whole-body model of the \textsc{Talos} robot, with single support time $T_\mathrm{ss}$ set to \SI{0.6}{s}, \SI{0.8}{s} and \SI{1}{s} from left to right. Each instance is run 40 times on every solver to produce a mean and standard deviation.\\
    Here, the Turboboost feature on the CPU was disabled and clock speed fixed to \SI{2200}{MHz}.}
    \label{fig:bench_std}
\end{figure}

We consider a whole-body trajectory optimization problem on a \textsc{Talos}~\cite{stasseTALOSNewHumanoid2017} humanoid robot with constrained 6D contacts. For this robot, the state and control dimensions are $n_x = 57$ and $n_u = 22$ respectively. The robot follows a user-defined contact sequence with feet references going smoothly from one contact to the next. State and controls are regularized towards an initial static half-sitting position and $u=0$ respectively. Single-foot support time $T_\text{ss}$ is set to 4 times double-support time $T_\text{ds}$. We consider three instances of the problem with different time horizons, encompassing two full steps of the robot. The problem is discretized using the semi-implicit Euler scheme with timestep $\Delta t =\SI{10}{ms}$. In \cref{fig:bench_std}, our proximal solver with various parallelization settings is compared against the feasibility-prone DDP from the \textsc{Crocoddyl} library~\cite{mastalliCrocoddylEfficientVersatile2020}.

\subsubsection{Constrained NMPC on \textsc{Talos}}

\begin{figure}[ht!]
    \centering
    \includegraphics[width=\linewidth]{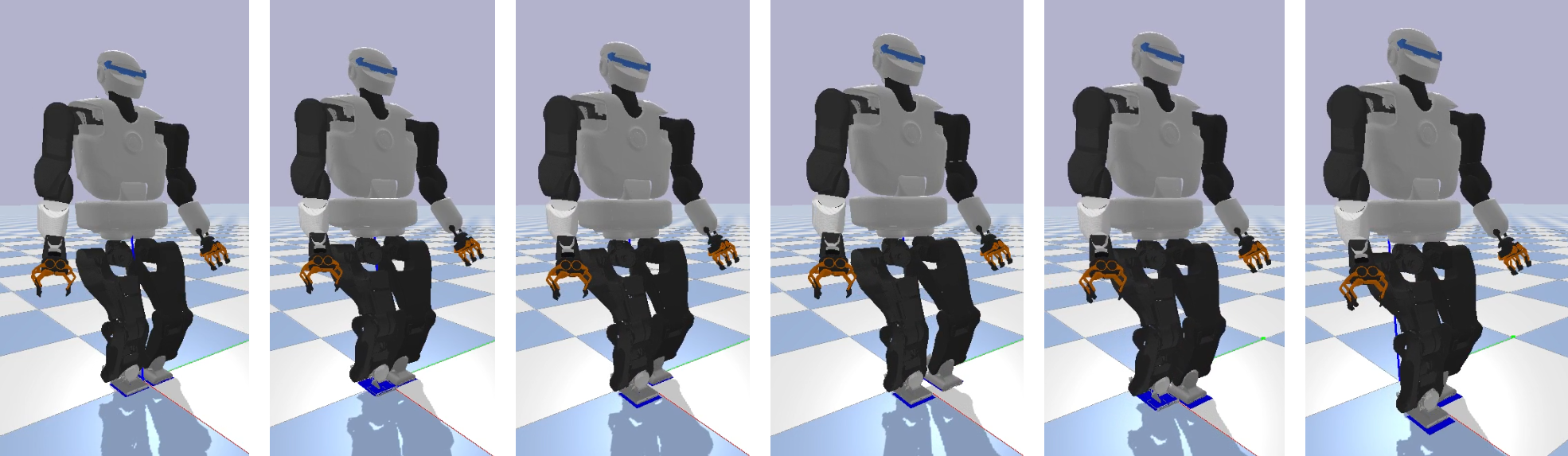}
    \caption{Snapshot of a PyBullet~\cite{coumansPyBulletPythonModule2016} simulation featuring \textsc{Talos} walking with pre-defined feet trajectories (blue rectangles). The Bullet simulation timestep is set to \SI{1}{ms}.}
    \label{fig:talos_snapshot}
\end{figure}

In this subsection, we leverage our proximal solver to perform whole-body nonlinear MPC on the humanoid robot \textsc{Talos} in simulation, similarly to what is achieved in~\cite{dantecWholeBodyModelPredictive2022} on real hardware. The problem remains the same as in the previous subsection, except that we add equality constraints at the end of each flying phase to ensure that foot altitude and 6D velocity are zero at impact time ($7$ constraints per foot). Horizon window is set to \SI{0.5}{s} with a timestep of \SI{10}{ms} for a total of $N=50$ steps. A snapshot of this experiment is displayed in \cref{fig:talos_snapshot}.

\begin{figure}[ht!]
    \centering
    \includegraphics[width=0.9\linewidth]{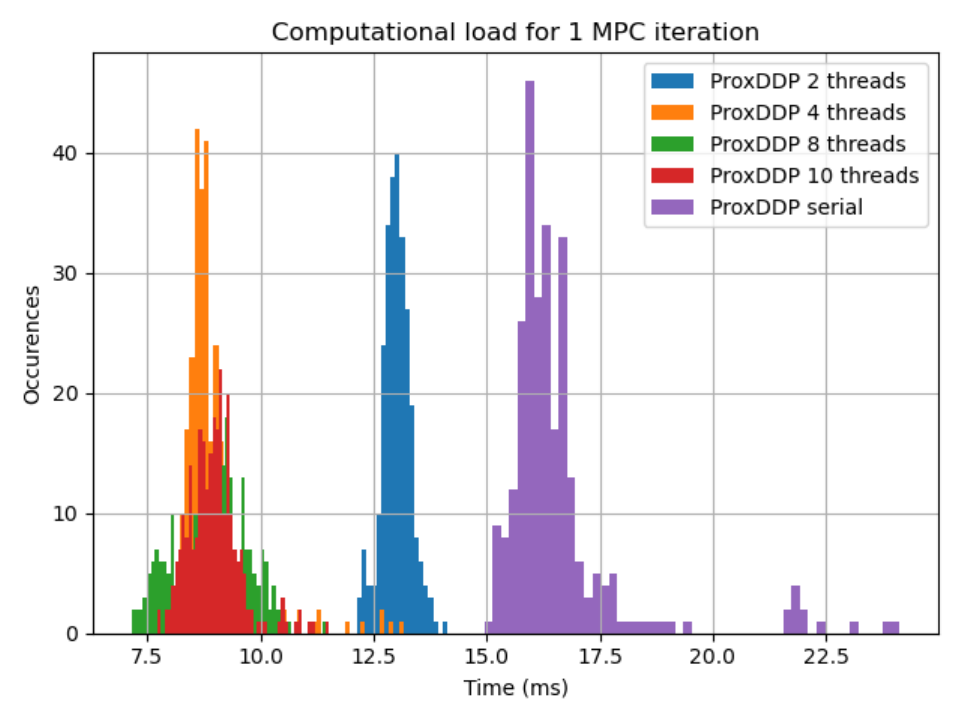}
    \caption{Comparison of the performances of parallel and serial proximal algorithms on the \textsc{Talos} walking MPC. The histogram shows the distribution of time per iteration for a two-step motion. The simulation was written using PyBullet, and the proximal solver was called through its Python API.}
    \label{fig:timing_MPC}
\end{figure}

In order to test the benefits of parallelization in the NMPC setting, the walking motion experiment has been run with varying numbers of threads. The timing results of this experiment are shown in \cref{fig:timing_MPC}. When using 4 threads, computation time is cut by a factor 2 with respect to the serial algorithm. However, above 4 threads, the speedup stagnates.

\subsubsection{Parallel NMPC on \textsc{Solo-12}}

To demonstrate the capabilities of our solver in a real-world setting, we integrated it inside a whole-body nonlinear MPC framework to control an actual torque-controlled quadruped robot, \textsc{Solo-12}, adapting the framework of~\citet{assirelliWholeBodyMPCFoot}.

The framework formulates an optimal control problem (OCP) with a state of dimension $n_x = 37$ ($12+12$ joint positions/velocities and $7+6$ base pose/velocity) and controls of size $n_u = 12$. Akin to the experiment on \textsc{Talos}, the robot follows a user-defined contact sequence, yet with no predefined reference foot trajectories. The horizon is set to \SI{0.96}{s}, with a \SI{12}{ms} timestep, resulting in a discrete-time horizon of $N=80$.

The above framework is divided into two parts, running on separate computers communicating via local Ethernet:
\begin{itemize}
    \item a high-level OCP solver runs on a powerful desktop,
    \item a lower-level, high-frequency controller on a laptop with the necessary drivers for real-time robot control. It interpolates the Riccati gains and feedforward controls between two MPC cycles, producing a reference torque and joint trajectory for the robot at \SI{1}{k\hertz}.
\end{itemize}

The time budget for solving the OCP is approximately 9~ms. The MPC has a frequency of 12~ms (same as the discretization step of the OCP), communication requires \SI{2}{ms} (round-trip), and \SI{1}{ms} is left for margin and additional computations. This limited budget, coupled with the problem's long horizon and dimensionality, justifies the need for parallelization.

The solve times for a single iteration of the OCP in the MPC loop were measured over 20,000 MPC cycles in the experiment. \Cref{tab:solo_walk:times} summarizes some statistics.
\begin{table}[ht!]
	\centering
	\begin{tabular}{@{}l|ccccc}
		\toprule
		\textbf{No. of threads} & 2 & 4 & 8 & 10 & 12 \\ \midrule
		Mean time (\SI{}{ms}) & 10.3 & 6.0 & 4.6 & 4.6 & 4.4 \\
		Std. dev (\SI{}{ms}) & 1.0 & 1.2 & 0.97 & 0.90 & 0.85 \\
		\bottomrule
	\end{tabular}
	\caption{Mean time (and standard deviation) for the MPC experiment on \textsc{Solo-12}. Statistics computed over 20,000 control cycles.}
	\label{tab:solo_walk:times}
\end{table}

\begin{figure}[t!]
    \centering
    \includegraphics[width=\linewidth]{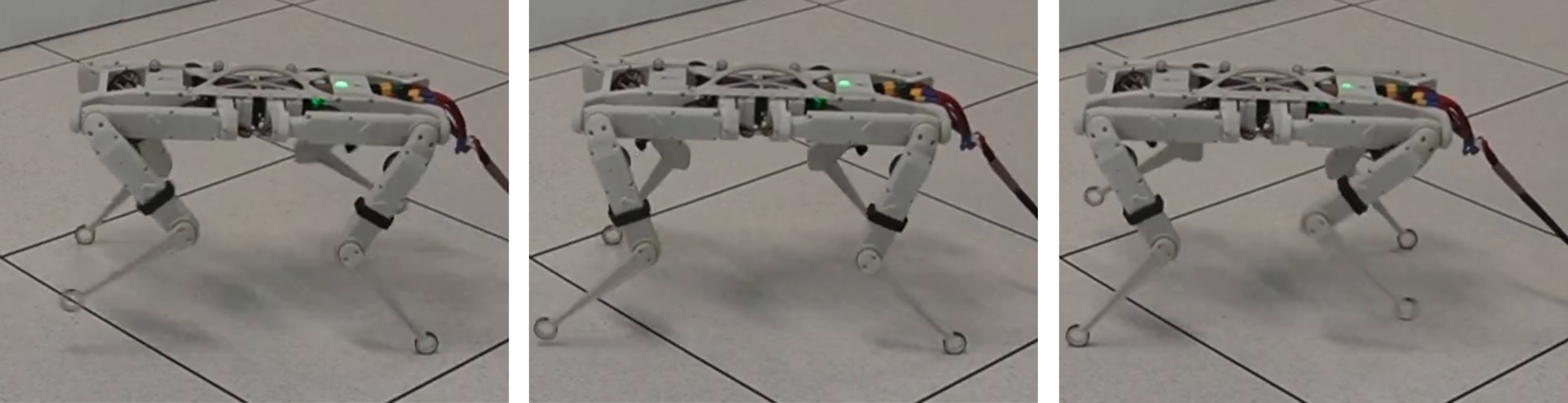}
    \caption{\textsc{Solo-12} walking on flat ground.}
    \label{fig:solo_walk}
\end{figure}

\noindent \textbf{Computer specifications:} Apple Mac Studio M1 Ultra desktop with 20 cores (16 P-cores and 4 E-cores), and Dell \textsc{xps} laptop with an Intel Core i7-10510U CPU.

\section{Conclusion}

In this paper, we have discussed the proximal-regularized LQ problem, as a subproblem in nonlinear MPC solvers, and have introduced a serial Riccati-like algorithm with a block-sparse resolution method which allows efficient resolution of this proximal iteration subproblem. Furthermore, we extended this method to solving parametric LQ problems and then leveraged this to formulate a parallel version of the initial algorithm.

We demonstrated that the parallel algorithm is able to handle complex tasks on high-dimensional robots with long horizons, even for real-time NMPC scenarios.
Still, while we are able to reach good overall execution times, the benchmarks suggest our implementation is not able to reach high parallel efficiency as of yet.
This suggests an avenue for further work on this implementation and further experimental validation and benchmarking.
Furthermore, the timings on the tested systems open the door to experimenting with more complex motions, assessing the benefits of either longer time horizons or higher-frequency MPC schemes.

\section*{Acknowledgements}

WJ thanks Ajay Sathya for his insights into block-matrix factorization in proximal settings.
This work was supported in part by the French government under the management of Agence Nationale de la Recherche (ANR) as part of the ”Investissements d'avenir” program, references ANR-19-P3IA-0001 (\textsc{PRAIRIE} 3IA Institute), ANR-19-p3IA-0004 (\textsc{ANITI} 3IA Institute) and ANR-22-CE33-0007 (INEXACT), the European project AGIMUS (Grant 101070165), the Louis Vuitton ENS Chair on Artificial Intelligence and the Casino ENS Chair on Algorithmic and Machine Learning.
Any opinions, findings, conclusions, or recommendations expressed in this material are those of the authors and do not necessarily reflect the views of the funding agencies.


\bibliographystyle{plainnat}
\bibliography{references,other-refs}

\begin{thebibliography}{52}
\providecommand{\natexlab}[1]{#1}
\providecommand{\url}[1]{\texttt{#1}}
\expandafter\ifx\csname urlstyle\endcsname\relax
  \providecommand{\doi}[1]{doi: #1}\else
  \providecommand{\doi}{doi: \begingroup \urlstyle{rm}\Url}\fi

\bibitem[Adabag et~al.(2023)Adabag, Atal, Gerard, and
  Plancher]{adabagMPCGPURealTimeNonlinear2023}
Emre Adabag, Miloni Atal, William Gerard, and Brian Plancher.
\newblock Mpcgpu: Real-time nonlinear model predictive control through
  preconditioned conjugate gradient on the gpu, September 2023.

\bibitem[Amos et~al.()Amos, Jimenez, Sacks, Boots, and
  Kolter]{amosDifferentiableMPCEndtoend}
Brandon Amos, Ivan Jimenez, Jacob Sacks, Byron Boots, and J~Zico Kolter.
\newblock Differentiable mpc for end-to-end planning and control.
\newblock page~12.

\bibitem[Assirelli et~al.()Assirelli, Risbourg, Lunardi, Flayols, and
  Mansard]{assirelliWholeBodyMPCFoot}
Alessandro Assirelli, Fanny Risbourg, Gianni Lunardi, Thomas Flayols, and
  Nicolas Mansard.
\newblock Whole-body mpc without foot references for the locomotion of an
  impedance-controlled robot.

\bibitem[Bertsekas(2019)]{bertsekas2019reinforcement}
Dimitri Bertsekas.
\newblock \emph{Reinforcement learning and optimal control}.
\newblock Athena Scientific, 2019.

\bibitem[Bounou et~al.(2023)Bounou, Ponce, and
  Carpentier]{bounouLeveragingProximalOptimization2023}
Oumayma Bounou, Jean Ponce, and Justin Carpentier.
\newblock Leveraging proximal optimization for differentiating optimal control
  solvers.
\newblock In \emph{IEEE 62nd Conference on Decision and Control (CDC)},
  Singapore, Singapore, December 2023.

\bibitem[Boyd()]{boydProximalAlgorithms}
Stephen Boyd.
\newblock Proximal algorithms.
\newblock page 113.

\bibitem[Boyd(2010)]{boydDistributedOptimizationStatistical2010}
Stephen Boyd.
\newblock Distributed optimization and statistical learning via the alternating
  direction method of multipliers.
\newblock \emph{Foundations and Trends{\textregistered} in Machine Learning},
  3\penalty0 (1):\penalty0 1--122, 2010.
\newblock ISSN 1935-8237, 1935-8245.
\newblock \doi{10.1561/2200000016}.

\bibitem[Bu and Plancher(2023)]{buSymmetricStairPreconditioning2023}
Xueyi Bu and Brian Plancher.
\newblock Symmetric stair preconditioning of linear systems for parallel
  trajectory optimization, September 2023.

\bibitem[Bunch and Kaufman(1977)]{bunchStableMethodsCalculating1977}
James Bunch and Linda Kaufman.
\newblock Some stable methods for calculating inertia and solving symmetric
  linear systems.
\newblock \emph{Mathematics of Computation - Math. Comput.}, 31:\penalty0
  163--163, January 1977.
\newblock \doi{10.1090/S0025-5718-1977-0428694-0}.

\bibitem[Carpentier et~al.(2019)Carpentier, Saurel, Buondonno, Mirabel,
  Lamiraux, Stasse, and Mansard]{carpentierPinocchioLibraryFast2019}
Justin Carpentier, Guilhem Saurel, Gabriele Buondonno, Joseph Mirabel, Florent
  Lamiraux, Olivier Stasse, and Nicolas Mansard.
\newblock The pinocchio c++ library -- a fast and flexible implementation of
  rigid body dynamics algorithms and their analytical derivatives.
\newblock In \emph{IEEE International Symposium on System Integrations (SII)},
  2019.

\bibitem[Carpentier et~al.(2021)Carpentier, Budhiraja, and
  Mansard]{carpentierProximalSparseResolution2021}
Justin Carpentier, Rohan Budhiraja, and Nicolas Mansard.
\newblock Proximal and sparse resolution of constrained dynamic equations.
\newblock In \emph{Robotics: Science and Systems XVII}. {Robotics: Science and
  Systems Foundation}, July 2021.
\newblock ISBN 978-0-9923747-7-8.
\newblock \doi{10.15607/RSS.2021.XVII.017}.

\bibitem[Chandra et~al.(2001)Chandra, Dagum, Kohr, Menon, Maydan, and
  McDonald]{chandra2001parallel}
Rohit Chandra, Leo Dagum, David Kohr, Ramesh Menon, Dror Maydan, and Jeff
  McDonald.
\newblock \emph{Parallel programming in OpenMP}.
\newblock Morgan kaufmann, 2001.

\bibitem[Chen et~al.(2008)Chen, Davis, Hager, and
  Rajamanickam]{chenAlgorithm887CHOLMOD2008}
Yanqing Chen, Timothy~A. Davis, William~W. Hager, and Sivasankaran
  Rajamanickam.
\newblock Algorithm 887: Cholmod, supernodal sparse cholesky factorization and
  update/downdate.
\newblock \emph{ACM Transactions on Mathematical Software}, 35\penalty0
  (3):\penalty0 1--14, October 2008.
\newblock ISSN 0098-3500, 1557-7295.
\newblock \doi{10.1145/1391989.1391995}.

\bibitem[Coumans and Bai(2016/2020)]{coumansPyBulletPythonModule2016}
Erwin Coumans and Yunfei Bai.
\newblock \emph{PyBullet, a Python Module for Physics Simulation for Games,
  Robotics and Machine Learning}.
\newblock 2016/2020.

\bibitem[Dantec et~al.(2022{\natexlab{a}})Dantec, Naveau, Fernbach, Villa,
  Saurel, Stasse, Taix, and Mansard]{dantecWholeBodyModelPredictive2022}
Ewen Dantec, Maximilien Naveau, Pierre Fernbach, Nahuel Villa, Guilhem Saurel,
  Olivier Stasse, Michel Taix, and Nicolas Mansard.
\newblock Whole-body model predictive control for biped locomotion on a
  torque-controlled humanoid robot.
\newblock In \emph{2022 IEEE-RAS 21st International Conference on Humanoid
  Robots (Humanoids)}, pages 638--644, November 2022{\natexlab{a}}.
\newblock \doi{10.1109/Humanoids53995.2022.10000129}.

\bibitem[Dantec et~al.(2022{\natexlab{b}})Dantec, Ta{\"i}x, and
  Mansard]{dantecFirstOrderApproximation2022}
Ewen Dantec, Michel Ta{\"i}x, and Nicolas Mansard.
\newblock First order approximation of model predictive control solutions for
  high frequency feedback.
\newblock \emph{IEEE Robotics and Automation Letters}, 7\penalty0 (2):\penalty0
  4448--4455, April 2022{\natexlab{b}}.
\newblock ISSN 2377-3766.
\newblock \doi{10.1109/LRA.2022.3149573}.

\bibitem[Davis(2005)]{davisAlgorithm849Concise2005}
Timothy~A. Davis.
\newblock Algorithm 849: A concise sparse cholesky factorization package.
\newblock \emph{ACM Transactions on Mathematical Software}, 31\penalty0
  (4):\penalty0 587--591, December 2005.
\newblock ISSN 0098-3500.
\newblock \doi{10.1145/1114268.1114277}.

\bibitem[Deng and Ohtsuka(2018)]{dengHighlyParallelizableNewtontype2018}
Haoyang Deng and Toshiyuki Ohtsuka.
\newblock A highly parallelizable newton-type method for nonlinear model
  predictive control.
\newblock \emph{IFAC-PapersOnLine}, 51:\penalty0 349--355, January 2018.
\newblock \doi{10.1016/j.ifacol.2018.11.058}.

\bibitem[Diehl et~al.(2006)Diehl, Bock, Diedam, and
  Wieber]{diehlFastDirectMultiple2006}
M.~Diehl, H.G. Bock, H.~Diedam, and P.-B. Wieber.
\newblock Fast direct multiple shooting algorithms for optimal robot control.
\newblock In Moritz Diehl and Katja Mombaur, editors, \emph{Fast Motions in
  Biomechanics and Robotics}, volume 340, pages 65--93. Springer Berlin
  Heidelberg, Berlin, Heidelberg, 2006.
\newblock ISBN 978-3-540-36118-3.
\newblock \doi{10.1007/978-3-540-36119-0_4}.

\bibitem[Dunn and Bertsekas(1989)]{dunnEfficientDynamicProgramming1989}
J.~C. Dunn and D.~P. Bertsekas.
\newblock Efficient dynamic programming implementations of newton's method for
  unconstrained optimal control problems.
\newblock \emph{Journal of Optimization Theory and Applications}, 63\penalty0
  (1):\penalty0 23--38, October 1989.
\newblock ISSN 1573-2878.
\newblock \doi{10.1007/BF00940728}.

\bibitem[Farshidian et~al.()]{ocs2}
Farbod Farshidian et~al.
\newblock {OCS2}: An open source library for optimal control of switched
  systems.
\newblock [Online]. Available: \url{https://github.com/leggedrobotics/ocs2}.

\bibitem[Frison()]{frisonAlgorithmsMethodsHighPerformance}
Gianluca Frison.
\newblock Algorithms and methods for high-performance model predictive control.
\newblock page 345.

\bibitem[Giftthaler and Buchli(2017)]{giftthalerProjectionApproachEquality2017}
Markus Giftthaler and Jonas Buchli.
\newblock A projection approach to equality constrained iterative linear
  quadratic optimal control.
\newblock \emph{2017 IEEE-RAS 17th International Conference on Humanoid
  Robotics (Humanoids)}, pages 61--66, November 2017.
\newblock \doi{10.1109/HUMANOIDS.2017.8239538}.

\bibitem[Giftthaler et~al.(2018)Giftthaler, Neunert, St{\"a}uble, Buchli, and
  Diehl]{giftthalerFamilyIterativeGaussNewton2018}
Markus Giftthaler, Michael Neunert, M.~St{\"a}uble, J.~Buchli, and M.~Diehl.
\newblock A family of iterative gauss-newton shooting methods for nonlinear
  optimal control.
\newblock \emph{2018 IEEE/RSJ International Conference on Intelligent Robots
  and Systems (IROS)}, 2018.
\newblock \doi{10.1109/IROS.2018.8593840}.

\bibitem[Guennebaud et~al.(2010)Guennebaud, Jacob, et~al.]{eigenweb}
Gaël Guennebaud, Benoît Jacob, et~al.
\newblock Eigen v3.
\newblock http://eigen.tuxfamily.org, 2010.

\bibitem[Howell et~al.(2019)Howell, Jackson, and
  Manchester]{howellALTROFastSolver2019}
Taylor~A. Howell, Brian~E. Jackson, and Zachary Manchester.
\newblock Altro: A fast solver for constrained trajectory optimization.
\newblock In \emph{2019 IEEE/RSJ International Conference on Intelligent Robots
  and Systems (IROS)}, pages 7674--7679, Macau, China, November 2019. IEEE.
\newblock ISBN 978-1-72814-004-9.
\newblock \doi{10.1109/IROS40897.2019.8967788}.

\bibitem[Jacobson and Mayne(1970)]{jacobsonDifferentialDynamicProgramming1970}
David~H. Jacobson and David~Q. Mayne.
\newblock \emph{Differential Dynamic Programming}.
\newblock American Elsevier Publishing Company, 1970.
\newblock ISBN 978-0-444-00070-5.

\bibitem[Jallet et~al.(2022{\natexlab{a}})Jallet, Bambade, Mansard, and
  Carpentier]{jalletConstrainedDifferentialDynamic2022}
Wilson Jallet, Antoine Bambade, Nicolas Mansard, and Justin Carpentier.
\newblock Constrained differential dynamic programming: A primal-dual augmented
  lagrangian approach.
\newblock In \emph{2022 IEEE/RSJ International Conference on Intelligent Robots
  and Systems}, Kyoto, Japan, October 2022{\natexlab{a}}.
\newblock \doi{10.1109/IROS47612.2022.9981586}.

\bibitem[Jallet et~al.(2022{\natexlab{b}})Jallet, Mansard, and
  Carpentier]{jalletImplicitDifferentialDynamic2022}
Wilson Jallet, Nicolas Mansard, and Justin Carpentier.
\newblock Implicit differential dynamic programming.
\newblock In \emph{2022 International Conference on Robotics and Automation
  (ICRA)}, Philadelphia, United States, May 2022{\natexlab{b}}. {IEEE Robotics
  and Automation Society}.
\newblock \doi{10.1109/ICRA46639.2022.9811647}.

\bibitem[Jallet et~al.(2023)Jallet, Bambade, Arlaud, {El-Kazdadi}, Mansard, and
  Carpentier]{jalletPROXDDPProximalConstrained2023}
Wilson Jallet, Antoine Bambade, Etienne Arlaud, Sarah {El-Kazdadi}, Nicolas
  Mansard, and Justin Carpentier.
\newblock Proxddp: Proximal constrained trajectory optimization, 2023.

\bibitem[Jordana et~al.(2023)Jordana, Kleff, Meduri, Carpentier, Mansard, and
  Righetti]{jordanaStagewiseImplementationsSequential2023}
Armand Jordana, S{\'e}bastien Kleff, Avadesh Meduri, Justin Carpentier, Nicolas
  Mansard, and Ludovic Righetti.
\newblock Stagewise implementations of sequential quadratic programming for
  model-predictive control, December 2023.

\bibitem[Kazdadi et~al.(2021)Kazdadi, Carpentier, and
  Ponce]{kazdadiEqualityConstrainedDifferential2021}
Sarah Kazdadi, Justin Carpentier, and Jean Ponce.
\newblock Equality constrained differential dynamic programming.
\newblock In \emph{ICRA 2021 - IEEE International Conference on Robotics and
  Automation}, May 2021.

\bibitem[Laine and
  Tomlin(2019{\natexlab{a}})]{laineEfficientComputationFeedback2019}
Forrest Laine and Claire Tomlin.
\newblock Efficient computation of feedback control for equality-constrained
  lqr.
\newblock In \emph{2019 International Conference on Robotics and Automation
  (ICRA)}, pages 6748--6754, May 2019{\natexlab{a}}.
\newblock \doi{10.1109/ICRA.2019.8793566}.

\bibitem[Laine and
  Tomlin(2019{\natexlab{b}})]{laineParallelizingLQRComputation2019}
Forrest Laine and Claire Tomlin.
\newblock Parallelizing lqr computation through endpoint-explicit riccati
  recursion.
\newblock In \emph{2019 IEEE 58th Conference on Decision and Control (CDC)},
  pages 1395--1402, December 2019{\natexlab{b}}.
\newblock \doi{10.1109/CDC40024.2019.9029974}.

\bibitem[Li et~al.(2023)Li, Yu, Zhang, and
  Wensing]{liUnifiedPerspectiveMultiple2023}
He~Li, Wenhao Yu, Tingnan Zhang, and Patrick~M. Wensing.
\newblock A unified perspective on multiple shooting in differential dynamic
  programming.
\newblock In \emph{2023 IEEE/RSJ International Conference on Intelligent Robots
  and Systems (IROS)}, pages 9978--9985, October 2023.
\newblock \doi{10.1109/IROS55552.2023.10342217}.

\bibitem[Mastalli et~al.(2020)Mastalli, Budhiraja, Merkt, Saurel, Hammoud,
  Naveau, Carpentier, Righetti, Vijayakumar, and
  Mansard]{mastalliCrocoddylEfficientVersatile2020}
Carlos Mastalli, Rohan Budhiraja, Wolfgang Merkt, Guilhem Saurel, Bilal
  Hammoud, Maximilien Naveau, Justin Carpentier, Ludovic Righetti, Sethu
  Vijayakumar, and Nicolas Mansard.
\newblock Crocoddyl: An efficient and versatile framework for multi-contact
  optimal control.
\newblock \emph{2020 IEEE International Conference on Robotics and Automation
  (ICRA)}, pages 2536--2542, May 2020.
\newblock \doi{10.1109/ICRA40945.2020.9196673}.

\bibitem[Mayne(1966)]{mayneSecondorderGradientMethod1966}
David Mayne.
\newblock A second-order gradient method for determining optimal trajectories
  of non-linear discrete-time systems.
\newblock \emph{International Journal of Control}, 3\penalty0 (1):\penalty0
  85--95, January 1966.
\newblock ISSN 0020-7179.
\newblock \doi{10.1080/00207176608921369}.

\bibitem[Nielsen and Axehill(2014)]{nielsenLogParallelAlgorithm2014}
Isak Nielsen and Daniel Axehill.
\newblock An {\^o} (log n) parallel algorithm for newton step computation in
  model predictive control.
\newblock \emph{IFAC Proceedings Volumes}, 47\penalty0 (3):\penalty0
  10505--10511, January 2014.
\newblock ISSN 1474-6670.
\newblock \doi{10.3182/20140824-6-ZA-1003.01577}.

\bibitem[Nielsen and Axehill(2015)]{nielsenParallelStructureExploiting2015}
Isak Nielsen and Daniel Axehill.
\newblock A parallel structure exploiting factorization algorithm with
  applications to model predictive control.
\newblock In \emph{2015 54th IEEE Conference on Decision and Control (CDC)},
  pages 3932--3938, December 2015.
\newblock \doi{10.1109/CDC.2015.7402830}.

\bibitem[Nocedal and Wright(2006)]{nocedalNumericalOptimization2006}
Jorge Nocedal and Stephen~J. Wright.
\newblock \emph{Numerical Optimization}.
\newblock Springer Series in Operations Research. Springer, New York, 2nd ed
  edition, 2006.
\newblock ISBN 978-0-387-30303-1.

\bibitem[Oshin et~al.(2022)Oshin, Houghton, Acheson, Gregory, and
  Theodorou]{oshinParameterizedDifferentialDynamic2022}
Alex Oshin, Matthew~D Houghton, Michael~J. Acheson, Irene~M. Gregory, and
  Evangelos Theodorou.
\newblock Parameterized differential dynamic programming.
\newblock In \emph{Robotics: Science and Systems XVIII}. {Robotics: Science and
  Systems Foundation}, June 2022.
\newblock ISBN 978-0-9923747-8-5.
\newblock \doi{10.15607/RSS.2022.XVIII.046}.

\bibitem[Plancher and
  Kuindersma(2020)]{plancherPerformanceAnalysisParallel2020}
Brian Plancher and Scott Kuindersma.
\newblock A performance analysis of parallel differential dynamic programming
  on a gpu.
\newblock In Marco Morales, Lydia Tapia, Gildardo {S{\'a}nchez-Ante}, and Seth
  Hutchinson, editors, \emph{Algorithmic Foundations of Robotics XIII},
  volume~14, pages 656--672. Springer International Publishing, Cham, 2020.
\newblock ISBN 978-3-030-44050-3 978-3-030-44051-0.
\newblock \doi{10.1007/978-3-030-44051-0_38}.

\bibitem[Rockafellar(1976)]{rockafellarAugmentedLagrangiansApplications1976}
R.~T. Rockafellar.
\newblock Augmented lagrangians and applications of the proximal point
  algorithm in convex programming.
\newblock \emph{Mathematics of Operations Research}, 1\penalty0 (2):\penalty0
  97--116, 1976.
\newblock ISSN 0364-765X.

\bibitem[Schenk et~al.(2001)Schenk, G{\"a}rtner, Fichtner, and
  Stricker]{schenkPARDISOHighperformanceSerial2001}
Olaf Schenk, Klaus G{\"a}rtner, Wolfgang Fichtner, and Andreas Stricker.
\newblock Pardiso: A high-performance serial and parallel sparse linear solver
  in semiconductor device simulation.
\newblock \emph{Future Generation Computer Systems}, 18\penalty0 (1):\penalty0
  69--78, September 2001.
\newblock ISSN 0167-739X.
\newblock \doi{10.1016/S0167-739X(00)00076-5}.

\bibitem[Srinivasan and Todorov(2017)]{srinivasanGraphicalNewton2017}
Akshay Srinivasan and Emanuel Todorov.
\newblock Graphical newton.
\newblock 2017.

\bibitem[Srinivasan and Todorov(2021)]{srinivasanComputingNewtonstepFaster2021}
Akshay Srinivasan and Emanuel Todorov.
\newblock Computing the newton-step faster than hessian accumulation.
\newblock \emph{arXiv:2108.01219 [cs, eess, math]}, August 2021.

\bibitem[Stasse et~al.(2017)Stasse, Flayols, Budhiraja, {Giraud-Esclasse},
  Carpentier, Mirabel, Del~Prete, Sou{\`e}res, Mansard, Lamiraux, Laumond,
  Marchionni, Tome, and Ferro]{stasseTALOSNewHumanoid2017}
O.~Stasse, T.~Flayols, R.~Budhiraja, K.~{Giraud-Esclasse}, J.~Carpentier,
  J.~Mirabel, A.~Del~Prete, P.~Sou{\`e}res, N.~Mansard, F.~Lamiraux, J.-P.
  Laumond, L.~Marchionni, H.~Tome, and F.~Ferro.
\newblock Talos: A new humanoid research platform targeted for industrial
  applications.
\newblock In \emph{2017 IEEE-RAS 17th International Conference on Humanoid
  Robotics (Humanoids)}, pages 689--695, November 2017.
\newblock \doi{10.1109/HUMANOIDS.2017.8246947}.

\bibitem[Stathopoulos et~al.(2013)Stathopoulos, Keviczky, and
  Wang]{stathopoulosHierarchicalTimesplittingApproach2013}
Georgios Stathopoulos, Tamas Keviczky, and Yang Wang.
\newblock A hierarchical time-splitting approach for solving finite-time
  optimal control problems.
\newblock \emph{2013 European Control Conference (ECC)}, pages 3089--3094, July
  2013.
\newblock \doi{10.23919/ECC.2013.6669702}.

\bibitem[Tassa et~al.(2012)Tassa, Erez, and
  Todorov]{tassaSynthesisStabilizationComplex2012}
Yuval Tassa, Tom Erez, and Emanuel Todorov.
\newblock Synthesis and stabilization of complex behaviors through online
  trajectory optimization.
\newblock In \emph{Proceedings of the IEEE/RSJ International Conference on
  Intelligent Robots and Systems.}, pages 4906--4913, October 2012.
\newblock ISBN 978-1-4673-1737-5.
\newblock \doi{10.1109/IROS.2012.6386025}.

\bibitem[Vanroye et~al.(2023)Vanroye, De~Schutter, and
  Decr{\'e}]{vanroyeGeneralizationRiccatiRecursion2023}
Lander Vanroye, Joris De~Schutter, and Wilm Decr{\'e}.
\newblock A generalization of the riccati recursion for equality-constrained
  linear quadratic optimal control, June 2023.

\bibitem[Wright(1990)]{wrightSolutionDiscretetimeOptimal1990}
Stephen~J Wright.
\newblock Solution of discrete-time optimal control problems on parallel
  computers.
\newblock \emph{Parallel Computing}, 16\penalty0 (2-3):\penalty0 221--237,
  December 1990.
\newblock ISSN 01678191.
\newblock \doi{10.1016/0167-8191(90)90060-M}.

\bibitem[Wright(1991)]{wrightPartitionedDynamicProgramming1991}
Stephen~J. Wright.
\newblock Partitioned dynamic programming for optimal control.
\newblock \emph{SIAM Journal on Optimization}, 1\penalty0 (4):\penalty0
  620--642, November 1991.
\newblock ISSN 1052-6234.
\newblock \doi{10.1137/0801037}.

\end{thebibliography}

\appendix

\subsection{Refresher on Riccati recursion}
\label{appx:base_riccati}

In this appendix, we provide a self-contained and concise refresher on Riccati recursion.
We will consider an unconstrained, unregularized linear-quadratic problem
\begin{subequations}\label{eq:lqr:unconstrained}
	\begin{alignat}{3}
		\min_{\bmx, \bmu} &\ J(\bmx, \bmu) \defeq
		\sum_{t=0}^{N-1} \ell_t(x_t, u_t) + \ell_N(x_N) \\
		\subjto
		&x_{t+1} = A_tx_t + B_tu_t + f_t \\
		& x_0 = x^0
	\end{alignat}
\end{subequations}
where the cost functions $\ell_t$ and $\ell_N$ are the same as \eqref{eq:lqr:problem:costs}, and $x^0 \in \RR^{n_x}$ is the problem's initial data.
Compared to problem \eqref{eq:lqr:problem}, this problem has \emph{no contraints}, an \emph{explicit initial condition} and \emph{explicit system dynamics}.

\noindent\textbf{Optimality conditions.}
The KKT conditions for \eqref{eq:lqr:unconstrained} read as follow: there exist co-states $\{\lambda_t\}_{0\leq t\leq N}$ such that
\begin{subequations}
\begin{align}
	\label{eq:unconstrained:opt:term}
	\lambda_N &= Q_Nx_N + q_N, \\
	0 &= x^0 - x_0
\end{align}
as well as, for $0\leq t < N$,
\begin{align}\label{eq:unconstrained:opt:adj}
	Q_tx_t + S_t u_t + q_t + A_t^\top \lambda_{t+1} &= \lambda_t, \\
	\label{eq:unconstrained:opt:u}
	S_t^\top x_t + R_tu_t + r_t + B_t^\top \lambda_{t+1} &= 0 \\
	\label{eq:unconstrained:opt:dyn}
	A_tx_t + B_tu_t + f_t - x_{t+1} &= 0.
\end{align}
\end{subequations}

\noindent\textbf{Recursion.}
The recursion hypothesis is as follows: there exists a semidefinite positive \emph{cost-to-go} matrix $P_t$ and vector $p_t$ such that
\begin{equation}
	\lambda_t = P_tx_t + p_t.
\end{equation}
This is true for $t=N$ owing to the terminal optimality condition \eqref{eq:unconstrained:opt:term}.

Now, let $0\leq t<N$ such that the recursion hypothesis is true for $t+1$, i.e. there exist $(P_{t+1}, p_{t+1})$ such that $\lambda_{t+1}=P_{t+1}x_{t+1}+p_{t+1}$. Then, plugging \eqref{eq:unconstrained:opt:dyn} into this leads to
\begin{equation*}
	\lambda_{t+1} = P_{t+1}(A_tx_t+B_tu_t+f_t) + p_{t+1}
\end{equation*}
which with \cref{eq:unconstrained:opt:adj,eq:unconstrained:opt:u} leads into
\begin{equation}\label{eq:unconstrained:opt:reduced}
	\begin{bmatrix}
		\hatQ_t & \hatS_t \\
		\hatS_t^\top & \hatR_t
	\end{bmatrix}
	\begin{bmatrix}
		x_t \\ u_t
	\end{bmatrix}
	+ \begin{bmatrix}
		\hatq_t \\ \hatr_t
	\end{bmatrix} = \begin{bmatrix}
	\lambda_t \\ 0
	\end{bmatrix},
\end{equation}
where we have denoted
\begin{subequations}
\begin{align}
	\hatQ_t &= Q_t + A_t^\top P_{t+1}A_t, \\
	\hatS_t &= S_t + A_t^\top P_{t+1}B_t, \\
	\hatR_t &= R_t + B_t^\top P_{t+1}B_t, \\
	\hatq_t &= q_t + A_t^\top (p_{t+1} + P_{t+1}f_t), \\
	\hatr_t &= r_t + B_t^\top (p_{t+1} + P_{t+1}f_t).
\end{align}
\end{subequations}

The following step in the Riccati recursion is to apply the Schur complement lemma to \eqref{eq:unconstrained:opt:reduced} (see~\cref{appx:schur}), requiring that $\hatR_t \succ 0$. This leads to
\begin{subequations}
\begin{equation}
	\lambda_t = (\hatQ_t - \hatS_t \hatR_t^{-1}\hatS_t^\top)x_t + \hatq_t - \hatS_t^\top \hatR_t^{-1}\hatr_t
\end{equation}
as well as the familiar linear feedback equation
\begin{equation}
	u_t = k_t + K_tx_t
\end{equation}
where we denote by $K_t=-\hatR_t^{-1}\hatS_t^\top$ and $k_t=-\hatR_t^{-1}\hatr_t$ the classic \emph{feedback} and \emph{feedforward} gains.
\end{subequations}
The recursion is then closed by setting
\begin{subequations}
\begin{equation}
	P_t \defeq \hatQ_t - \hatS_t\hatR_t^{-1}\hatS_t^\top \in \symmat{n_x}
\end{equation}
and
\begin{equation}
	p_t \defeq \hatq_t - \hatS_t\hatR_t^{-1}\hatr_t.
\end{equation}
\end{subequations}

To go further, we direct the reader to an application of this recursion in the context of nonlinear control for robotics in~\cite{tassaSynthesisStabilizationComplex2012}, which introduced the popular \emph{iterative LQR} (iLQR) algorithm.

\subsection{Schur complement lemma}
\label{appx:schur}

In this appendix, we will provide an overview of the Schur complement lemma.

\subsubsection{2x2 block linear system}

We consider a symmetric $2\times 2$ block linear system
\begin{equation}\label{eq:schur:2x2}
	\begin{bmatrix}
		G & J^\top \\
		J & -\Lambda
	\end{bmatrix}
	\begin{bmatrix}
		x \\ z
	\end{bmatrix} = -\begin{bmatrix}
		b \\ c
	\end{bmatrix}
\end{equation} 	
where $G$ (resp. $\Lambda$) is a $n\times n$ (resp. $m \times m$) symmetric matrix, $b \in \RR^n$, $c \in \RR^m$, and $J \in \RR^{m\times n}$.

Symmetric indefinite systems such as~\eqref{eq:schur:2x2} can be solved by straightforward $LU$, $LDL^\top$ or Bunch-Kaufman~\cite{bunchStableMethodsCalculating1977} decomposition. Specific methods exist for sparse matrices~\cite{davisAlgorithm849Concise2005,chenAlgorithm887CHOLMOD2008}.

Assuming nonsingular $\Lambda$, the Schur complement in $x$, solves $z$ as a function of $x$:
\begin{equation}
	z = \Lambda^{-1}(c + Jx)
\end{equation}
and leads to an equation in $x$:
\begin{equation}
	(G + J^\top\Lambda^{-1}J)x = -(b + J^\top\Lambda^{-1}c).
\end{equation}
Here, $G_\Lambda = G + J^\top\Lambda^{-1}J$ is called the \emph{Schur matrix}. If, furthermore, we have that $G\succeq 0$ and $\Lambda \succ 0$, then the Schur matrix is positive semidefinite.


In some applications (e.g. constrained forward dynamics~\cite{carpentierProximalSparseResolution2021}), $G$ exhibits sparsity and the the complement in $z$ is chosen instead, computing $G^{-1}$ -- this corresponds to the partial minimization $\min_x f(x,z)$.

\subsubsection{Variant: lemma for \cref{prop:param_value}}\label{appx:schur:2}

Consider the following parametric saddle-point problem:
\begin{equation}\label{eq:schur:saddlepoint}
	\psi^\star(z) = \min_x \max_y f(x, y, z)
\end{equation}
where $f$ is the following quadratic:
\begin{equation}\label{eq:schur:quad_f}
	f(x,y,z) = \frac{1}{2}
	\begin{bmatrix}
		x \\ y \\ z
	\end{bmatrix}
	\begin{bmatrix}
		Q & S & C^\top \\
		S^\top & R & D^\top \\
		C & D & H
	\end{bmatrix}
	\begin{bmatrix}
		x \\ y \\ z
	\end{bmatrix}
	+ q^\top x + r^\top y + c^\top z
\end{equation}
with appropriate assumptions on $(Q,S,R)$ to ensure that $f$ is convex-concave (namely, $Q\succeq 0$ and that $R\preceq 0$) and
\begin{equation}
	M =
	\begin{bmatrix}
		Q & S \\ S^\top &  R
	\end{bmatrix}
\end{equation}
is nonsingular.
The first-order conditions satisfied by a solution $(x^\star,y^\star)$ of \eqref{eq:schur:saddlepoint} read:
\begin{equation}
	\begin{bmatrix}
		Q & S & C^\top \\
		S^\top & R & D^\top
	\end{bmatrix}
	\begin{bmatrix}
		x^\star \\ y^\star \\ z
	\end{bmatrix} +
	\begin{bmatrix}
		q \\ r
	\end{bmatrix} = 0.
\end{equation}
By nonsingularity of $M$, it holds
\begin{enumerate}
	\item[(i)] that $(x^\star,y^\star)$ is linear in $z$ with
	\begin{equation}
		\begin{bmatrix}
			x^\star \\ y^\star
		\end{bmatrix} = -M^{-1}\left(
		\begin{bmatrix}
			q \\ r
		\end{bmatrix}
		+ \begin{bmatrix}
			C^\top \\ D^\top
		\end{bmatrix} z
		\right),
	\end{equation}
	and, as a consequence,
	\item[(ii)] that $\psi^\star$ is a quadratic function of $z$: there exist a matrix $W$ and vector $w$ such that
	\begin{equation}
		\psi^\star(z) = \frac{1}{2}  z^\top W z + w^\top z,
	\end{equation}
	and $(W, w)$ is given by
	\begin{equation}
		\begin{split}
			W &= H - \begin{bmatrix}
				C & D
			\end{bmatrix} M^{-1} \begin{bmatrix}
				C^\top \\ D^\top
			\end{bmatrix}, \\
			w &= c - \begin{bmatrix}
				C & D
			\end{bmatrix}M^{-1}\begin{bmatrix}
				q \\ r
			\end{bmatrix}.
		\end{split}
	\end{equation}
\end{enumerate}

\subsection{Thomas algorithm for block-tridiagonal matrices}

Consider a block-sparse linear system
\begin{equation}
    \mathbb{M}\begin{bmatrix}
        X_1 \\ \vdots \\ X_N
    \end{bmatrix}
    = \begin{bmatrix}
        C_1 \\ \vdots \\ C_N
    \end{bmatrix}
\end{equation}
where the $X_i$ and $C_i$ are $n_i\times m$ matrices, and
\begin{equation}
    \mathbb{M} = \begin{bmatrix}
        A_0 & B_1 & \\
        B_1^\top & A_1 & B_2 \\
                 & B_2^\top & A_2 & \ddots \\
                 &      & \ddots & & B_N\\
        &   &   &       B_N^\top & A_N
    \end{bmatrix}
\end{equation}
is a block-triadiagonal matrix, where $A_i \in \RR^{n_i\times n_i}$, \mbox{$B_i \in \RR^{n_{i-1} \times n_i}$}.

\def\sfU{\mathsf{U}}
\def\sfD{\mathsf{D}}

A quick derivation of the backward-forward algorithm stems by considering an appropriate block $UDU^\top$ factorization, where we prescribe
\[
    U = \begin{bmatrix}
        I & \sfU_1 \\
          & I &  \\
          && \ddots \\
          &&& I & \sfU_N
    \end{bmatrix},\
    D = \begin{bmatrix}
        \sfD_0 & \\ & \sfD_1 \\ & & \ddots \\ & & & \sfD_N
    \end{bmatrix}.
\]
Writing $\mathbb{M} = UDU^\top$, it appears that a sufficient and necessary condition is that $(U,D)$ satisfy
\begin{subequations}
\begin{align}
    A_N &= \sfD_N \\
    B_i &= \sfU_i\sfD_i, \ 1\leq i \leq N \\
    A_i &= \sfD_i + \sfU_{i+1}\sfD_{i+1}\sfU_{i+1}^\top, \ 0\leq i < N,
\end{align}
\end{subequations}
which determines $(U,D)$ uniquely. The second and third equations can be combined together as $\sfD_i = A_i - B_{i+1} \sfD_{i+1}^{-1} B_{i+1}^\top $.

Furthermore, we use $(U,D)$ to solve the initial system as $UDU^\top X = C$, which can be split up as $U^\top X = Z$ and $UDZ = C$. The second equation can be solved by working backwards (since $U$ is block upper-triangular):
\begin{subequations}
\begin{align}
    Z_N &= \sfD_N^{-1} C_N \\
    Z_{i} &= \sfD_i^{-1}(C_i - B_{i+1}Z_{i+1}), \ 0 \leq i < N
\end{align}
\end{subequations}
and then solve the first equation for $X$, working forwards:
\begin{subequations}
\begin{align}
    X_0 &= Z_0 \\
    X_{i+1} &= Z_{i+1} - \sfU_{i+1}^\top X_i, \ 0\leq i < N.
\end{align}
\end{subequations}

\subsection{Formulating LQ problems with cyclic constraints}
\label{sec:param:cyclic}

Assume the LQ problem to solve is similar to \eqref{eq:lqr:problem} with a cyclical constraint
\[
	x_N - x_0 = 0
\]
and, for simplicity, no other path constraints (meaning $n_c = 0$).
We introduce for this constraint an additional multiplier $\theta \in \RR^{n_x}$ as a parameter.
The full problem Lagrangian is
\[
    \scrL(\bmx, \bmu, \bmlam; \theta) = \scrL(\bmx, \bmu, \bmlam) + \theta^\top (x_N - x_0).
\]

The terminal stage Lagrangian becomes $\scrL_N(x_N,\theta) = \ell_N(x_N) + \theta^\top x_N$. In this setting, our previous derivations apply with with $\Phi_{N} = I$.
After condensing into the parameterized value function $\calE(x_0, \theta)$, we can solve for $(x_0,\theta)$ by solving the min-max problem
\begin{equation}
    \max_\theta \min_{x_0} \calE(x_0, \theta) - \theta^\top x_0
\end{equation}
leading to the system
\begin{equation}
    \begin{bmatrix}
        P_0 & \Lambda_0 - I \\ (\Lambda_0 - I)^\top & 
    \end{bmatrix}
    \begin{bmatrix}
        x_0 \\ \theta
    \end{bmatrix} = -\begin{bmatrix}
        p_0 \\ \sigma_0
    \end{bmatrix}.
\end{equation}

Other topologies such as directed acyclic graphs (DAG) have been explored in the literature~\cite{srinivasanGraphicalNewton2017,srinivasanComputingNewtonstepFaster2021}. Such a discussion would be out of the scope of this paper, but it is our view that several algorithms in this vein can be restated using parametric Lagrangians and by formulating partial min-max problems.

\subsection{Details for the block-sparse factorization in \Cref{sec:blocksparse}}
\label{app:blocksparse}
%
We present here the details of how we propose to solve \eqref{eq:riccati:big_stage_system}-\eqref{eq:riccati:stage_kkt} by exploiting its block-sparse structure, replacing the less efficient recursion initially proposed in~\cite{jalletConstrainedDifferentialDynamic2022}.
%
The first step is isolating the equations, in \eqref{eq:riccati:big_stage_system}, satisfied by the co-state and next state $(\lambda_2, x_2)$. They are:
\begin{equation}\label{eq:riccati:costate_nextstate}
    \begin{bNiceArray}[first-row]{cc|l}
        \RowStyle{\arrstyle}
        \lambda_2 & x_2 & \\
        -\mu I & E_1 & \bar{f}_1 + A_1x_1 + B_1u_1 \\
        E_1^\top & P_2 & p_2
    \end{bNiceArray}.
\end{equation}
This system could be solved by Schur complement in either variables $\lambda_2$ or $x_2$. One way requires computing a decomposition of $P_2 + \tfrac{1}{\mu}E_1^\top E_1$, which is numerically unstable if $\mu$ is small. The other way requires $P_2$ to be nonsingular, and computing the inverse of $\mu I + E_1P_2^{-1}E_1^\top$ -- although this would be more stable, assuming nonsingular $P_2$ can be a hurdle in practical applications (typically, if the terminal cost only applies to part of the state space such as joint velocities).

\subsubsection{Substitution by $E_2$}
We approach this another way, by assuming that the dynamics matrix $E_1$ is nonsingular.
This means the substitution $\check{x}_2 = -E_1x_2$ is well-defined, and that we can define $\check{P}_2 = E_1^{-T} P_2E_1^{-1} $ and $\check{p}_2 = -E_1^{-T}p_2$.

Then, the system \eqref{eq:riccati:costate_nextstate} is equivalent to
\begin{equation}
    \begin{bNiceArray}[first-row]{cc|l}
        \RowStyle{\arrstyle}
        \lambda_2 & \check{x}_2  \\
        -\mu I & -I & \bar{f}_1 + A_1x_1 + B_1u_1 \\
        -I      & \check{P}_2 & \check{p}_2
    \end{bNiceArray}
\end{equation}
which by substituting $ \check{x}_2 = \bar{f}_1 + A_1x_1 + B_1u_1 - \mu\lambda_2 $ reduces to the symmetric linear system
\begin{equation}
    (\mu \check{P}_2 + I)\lambda_2 = \check{p}_2 + \check{P}_2(\bar{f}_1 + A_1x_1 + B_1u_1).
\end{equation}
Denote $\Upsilon = I + \mu\check{P}_2 $, and set $\calV_2 = \Upsilon^{-1} \check{P}_2$ and $v_2 = \Upsilon^{-1} (\check{p}_2 + \check{P}_2\bar{f}_1) $. Then, it holds that
\begin{equation*}
    \lambda_2 = v_2 + \calV_2(A_1x_1 + B_1u_1).
\end{equation*}
Substitution into \eqref{eq:riccati:big_stage_system} yields equations for $(x_1, u_1, \nu_1)$:
\begin{equation}\label{eq:riccati:reduced_stage_kkt}
	\begin{bNiceArray}[first-row]{ccc|l}
		\RowStyle{\arrstyle}
		x_1 & u_1 & \nu_1 \\
        \hatQ_1     & \hatS_1 & C_{1}^\top & \hatq_1 + E_0^\top\lambda_1 \\
		\hatS_1^\top& \hatR_1 & D_{1}^\top & \hatr_1 \\
		C_{1}   & D_{1} & -\mu I & \bar{h}_1 \\
	\end{bNiceArray}
\end{equation}
where $\hatQ_1, \hatS_1, \hatR_1, \hatq_1, \hatr_1$ are given by the familiar equations:
\begin{equation}\label{eq:riccati:hatted_mat}
\begin{aligned}
    \hatQ_1 &= Q_1 + A_1^\top\calV_2 A_1, \quad &\hatq_1 &= q_1 + A_1^\top v_2, \\
    \hatR_1 &= R_1 + B_1^\top\calV_2 B_1,       &\hatr_1 &= r_1 + B_1^\top v_2, \\
    \hatS_1 &= S_1 + A_1^\top\calV_2 B_1.
\end{aligned}
\end{equation}

\subsubsection{Multiplier and next-state update}
The closed-loop multiplier update is
\begin{equation}
    \lambda_2 = v_2 + \calV_2B_1 k_1 + \calV_2(A_1+B_1K_1)x_1 = \omega_2 + \Omega_2x_1
\end{equation}
and the next state update is
\begin{equation}
\begin{aligned}
    x_2 &= -E_1^{-1}\check{x}_2, \\
    \check{x}_2 &= (\bar{f}_1 + B_1 k_1 - \mu\omega_2) + (A_1+B_1K_1 - \mu\Omega_2)x_1 \\
                &= a_1 + M_1x_1.
\end{aligned}
\end{equation}

\subsubsection{Final substitution}
Now, our goal is to eliminate $(u_1, \nu_1)$ from the system, expressing them as a function of $x_1$.
Denote $\widehat{\calK}_1$ the lower-right $2\times 2$ block
\begin{equation}
	\widehat{\calK}_1 = \begin{bmatrix}
		\hatR_1 & D_1^\top \\
		D_1		& -\mu I
	\end{bmatrix},
\end{equation}
which is a nonsingular symmetric matrix due to the dual regularization block $-\mu I$.
Then, by Schur complement, we obtain $(u_1,\nu_1)$ as a feedback
\begin{equation}\label{eq:riccati:feedback}
    \begin{bmatrix}
        u_1 \\ \nu_1
    \end{bmatrix}
    = -\widehat{\calK}_1^{-1}
    \left(\begin{bmatrix}
        \hatr_1 \\ \bar{h}_1
    \end{bmatrix} +
    \begin{bmatrix}
        \hatS_1^\top \\ C_{1}
    \end{bmatrix}x_1\right)
\end{equation}
which can be rewritten as
\begin{equation}
    u_1 = k_1 + K_1x_1,\ \nu_1 = \zeta_1 + Z_1x_1.
\end{equation}

Then, it holds that the cost-to-go matrix and gradient introduced in \eqref{eq:riccati:cost_to_go} also satisfy
\begin{subequations}
\begin{align}
    P_1 &= \hatQ_1 + \hatS_1 K_1 + C_{1}^\top Z_1 \\
    p_1 &= \hatq_1 + \hatS_1k_1 + C_{1}^\top\zeta_1.
\end{align}
\end{subequations}

The full block-sparse algorithm is summarized in \Cref{alg:blocksparse_riccati}.

\begin{algorithm}[ht!]
    \KwData{Cost and constraint matrices $Q_1,S_1,R_1,q_1,r_1, A_1,B_1,E_1,\bar{f}_1,C_1,D_1,\bar{h}_1$, cost-to-go matrix and vector $P_2, p_2$}
    Compute $E_1^{-1}$\tcp*{using, e.g., $LU$}
    $\check{P}_2 \leftarrow E_1^{-T}P_2E_1^{-1} $\;
    $\check{p}_2 \leftarrow -E_1^{-T}p_2 $\;
    $\Upsilon \leftarrow \mu \check{P}_2 + I$ \;
    $\calV_2 \leftarrow \Upsilon^{-1}\check{P}_2 $\;
    $v_2 \leftarrow \Upsilon^{-1}(\check{p}_2 + \check{P}_2\bar{f}_1)$\;
    Set $(\hatQ_1,\hatS_1,\hatR_1,\hatq_1,\hatr_1)$ according to \eqref{eq:riccati:hatted_mat}\;
    Set $\widehat{\calK}_1 \leftarrow \begin{bsmallmatrix}
        \hatR_1 & D_1^\top \\ D_1 & -\mu I \end{bsmallmatrix}$\;
    Compute $ \widehat{\calK}_1^{-1}$ (using, e.g., $\text{LDL}^\text{T}$)\;
    \tcp{Solve \eqref{eq:riccati:feedback}}
    $\begin{bsmallmatrix}k_1 \\ \zeta_1\end{bsmallmatrix} \leftarrow -\widehat{\calK}_1^{-1} \begin{bsmallmatrix}
        \hatr_1 \\ \bar{h}_1
    \end{bsmallmatrix}$\;
    $\begin{bsmallmatrix}K_1 \\ Z_1\end{bsmallmatrix} \leftarrow -\widehat{\calK}_1^{-1} \begin{bsmallmatrix}
        \hatS_1^\top \\ C_1
    \end{bsmallmatrix}$\;
    $P_1 \leftarrow \hatQ_1 + \hatS_1K_1 + C_1^\top Z_1$\;
    $p_1 \leftarrow \hatq_1 + \hatS_1k_1 + C_1^\top \zeta_1$\;

	\caption{Block-sparse factorization for the stage KKT equations}\label{alg:blocksparse_riccati}
\end{algorithm}

\subsection{Bound for the parallel algorithm's speedup}

In this appendix, we etch a derivation for the complexity of the serial and parallel backward passes of our algorithms~\ref{alg:serial_riccati}, \ref{alg:param_riccati} and \ref{alg:parallel_lqr}, which we will compare to find an upper-bound for the expected speedup.

\def\ComplFac{\ensuremath{C_\text{fac}}}
\def\ComplRhs{\ensuremath{C_\text{col}}}
\def\ComplParam{\ensuremath{C_\text{param}}}

Consider the stage system~\eqref{eq:riccati:stage_kkt} in the backward pass of~\cref{alg:serial_riccati}.
We denote by:
\begin{itemize}
	\item $\ComplFac = \calO((n_x+n_u+n_c)^3)$ the complexity of the factorization step
	\item $\ComplRhs = \calO((n_x+n_u+n_c)^2)$ the complexity of the right-hand side solve for a single column.
\end{itemize}

\def\Tbwd{\ensuremath{T^\text{b}}}

Then, the primal-dual gains are recovered with complexity \mbox{$(n_x+1)\ComplRhs$}.
Furthermore, for the parametric algorithm, solving the parameter gains $(K_t^\theta,Z_t^\theta,\Omega_{t+1}^\theta,M_t^\theta)$ in \eqref{eq:param:stage_kkt_param} for a parameter of dimension $n_\theta$ has complexity $\ComplParam = n_\theta \ComplRhs$.

The complexity of the backward pass in the serial algorithm~\ref{alg:serial_riccati} is thus
\begin{equation}
	\Tbwd_\text{serial} = N(\ComplFac + (n_x+1)\ComplRhs).
\end{equation}

Now, consider the parallel algorithm executed over $J+1$ ($J\geq 1$) legs. Each stage in the backward pass for legs $0\leq j < J$ has complexity $\ComplFac + (2n_x+1)\ComplRhs$ (the last leg still has standard stage complexity $\ComplFac + (n_x+1)\ComplRhs$).
Meanwhile, the consensus system \eqref{eq:parallel:condensed_sys} has complexity \mbox{$C_\text{cons} \sim \tfrac{2J+1}{3}n_x^3$} (assuming Cholesky decompositions with complexity $n^3/3$).
We assume an equal split across the $J$ legs, such that $i_{j+1}-i_j = N/(J+1)$.
The parallel time complexity for this backward pass is thus
\begin{equation}
	\Tbwd_\text{parallel} = \underbracket{\tfrac{N}{J+1}
		(\ComplFac + (2n_x+1)\ComplRhs)}_{\text{parallel section}} + C_\text{cons}.
\end{equation}

\begin{figure}[hb!]
	\centering
	\includegraphics[width=\linewidth]{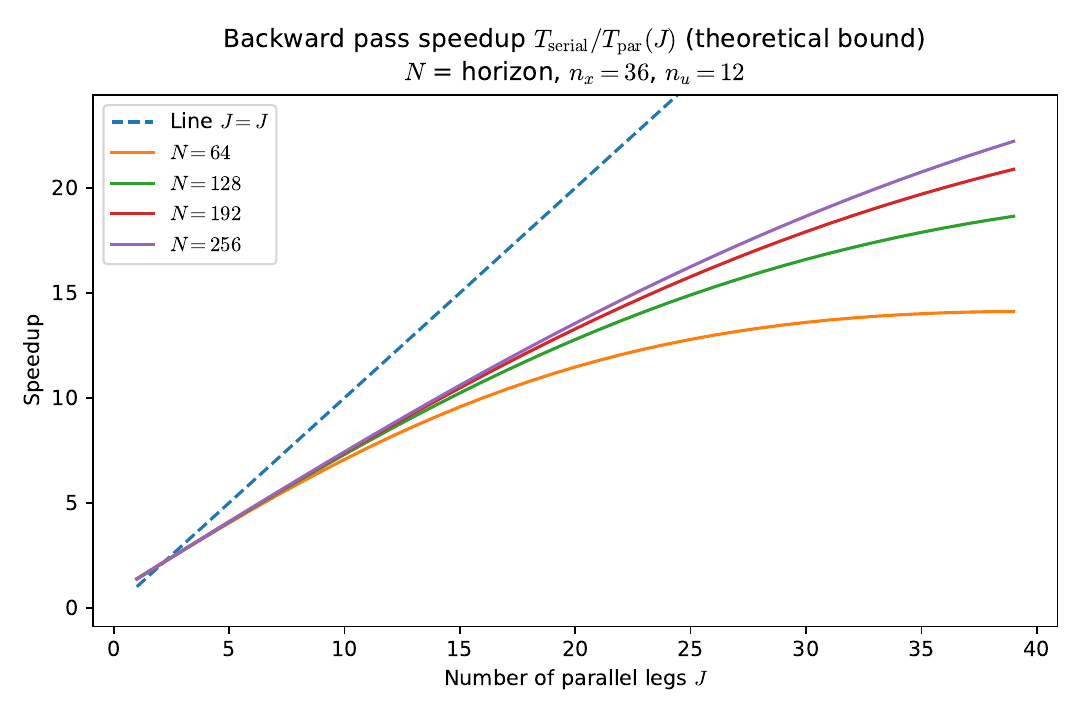}
	\caption{Theoretical speedup derived from the bounds, with varying problem horizon $N$ and number of parallel legs/processors $J$. The problem dimensions correspond to the Solo-12 system in \cref{sec:experiments}.}
	\label{fig:speedup_theoretical}
\end{figure}

We plot the speedup ratio $\Tbwd_\text{serial}/\Tbwd_\text{parallel}$ of serial to parallel execution times for a chosen dimension $(n_x,n_u,n_c)$ in \cref{fig:speedup_theoretical}. The graph shows early departure of this ratio from the ``perfect" speedup (which would be $\Tbwd_\text{parallel}=J\Tbwd_\text{serial}$).

\end{document}